\documentclass[10pt,twocolumn,letterpaper]{article}

\usepackage[pagenumbers]{iccv} 

\definecolor{iccvblue}{rgb}{0.21,0.49,0.74}
\usepackage[pagebackref,breaklinks,colorlinks,allcolors=iccvblue]{hyperref}

\def\paperID{7630} 
\def\confName{ICCV}
\def\confYear{2025}

\title{Learned JPEG Compression for DNN Vision}

\def\spaces{~~~~~~}
\author{Kaixiang Zheng\thanks{Equal contribution.}\spaces{} Ahmed H. Salamah\footnotemark[1]\spaces{}Siyu Chen\footnotemark[1]\spaces{}En-Hui Yang\\
Department of Electrical and Computer Engineering, University of Waterloo\\{\tt\small \{k56zheng, ahamsalamah, s875chen, ehyang\}@uwaterloo.ca}
}

\begin{document}
\maketitle
\begin{abstract}

JPEG, a lossy image compression technique designed for human viewers, has maintained its dominance for decades. However, in the era of artificial intelligence (AI), a substantial portion of image data, often compressed by JPEG, is and will continue to be consumed by deep neural networks (DNNs) instead of humans, thus creating a need to optimize JPEG for DNN inference performance. To this end, we propose learned JPEG compression for DNN vision (J4D), a novel training framework for determining JPEG encoding parameters to minimize compression rate while maximizing DNN inference performance. The major challenge of solving this optimization problem lies in representing the JPEG codec and compression rate in closed form. By incorporating a differentiable soft quantizer based on a probabilistic quantization scheme, we not only obtain a differentiable proxy for the JPEG codec, but are also able to compute the entropy of the coded source analytically, which is a close estimate of the actual compression rate. Equipped with both the differentiable JPEG codec and the information-theoretic rate estimator, we are then able to solve the aforementioned optimization problem with backpropagation. After training, the learned encoding parameters will be subsequently used in actual JPEG encoding based on probabilistic quantization. Extensive experimental results across multiple datasets and DNN architectures demonstrate that J4D consistently and significantly outperforms the default JPEG and other competitive JPEG codecs optimized for DNNs. Notably, compared to the default JPEG, J4D achieves an increase in accuracy by as much as 11.60\% at the same rate, or a reduction of compression rate up to 80.05\% at the same accuracy. Additionally, with the help of J4D, we show the potential to design universal JPEG encoding parameters for various DNN architectures for the first time.

\end{abstract}
    
\section{Introduction} \label{sec:Introduction}

\begin{figure}[t]
    \centering
    \includegraphics[width=\linewidth]{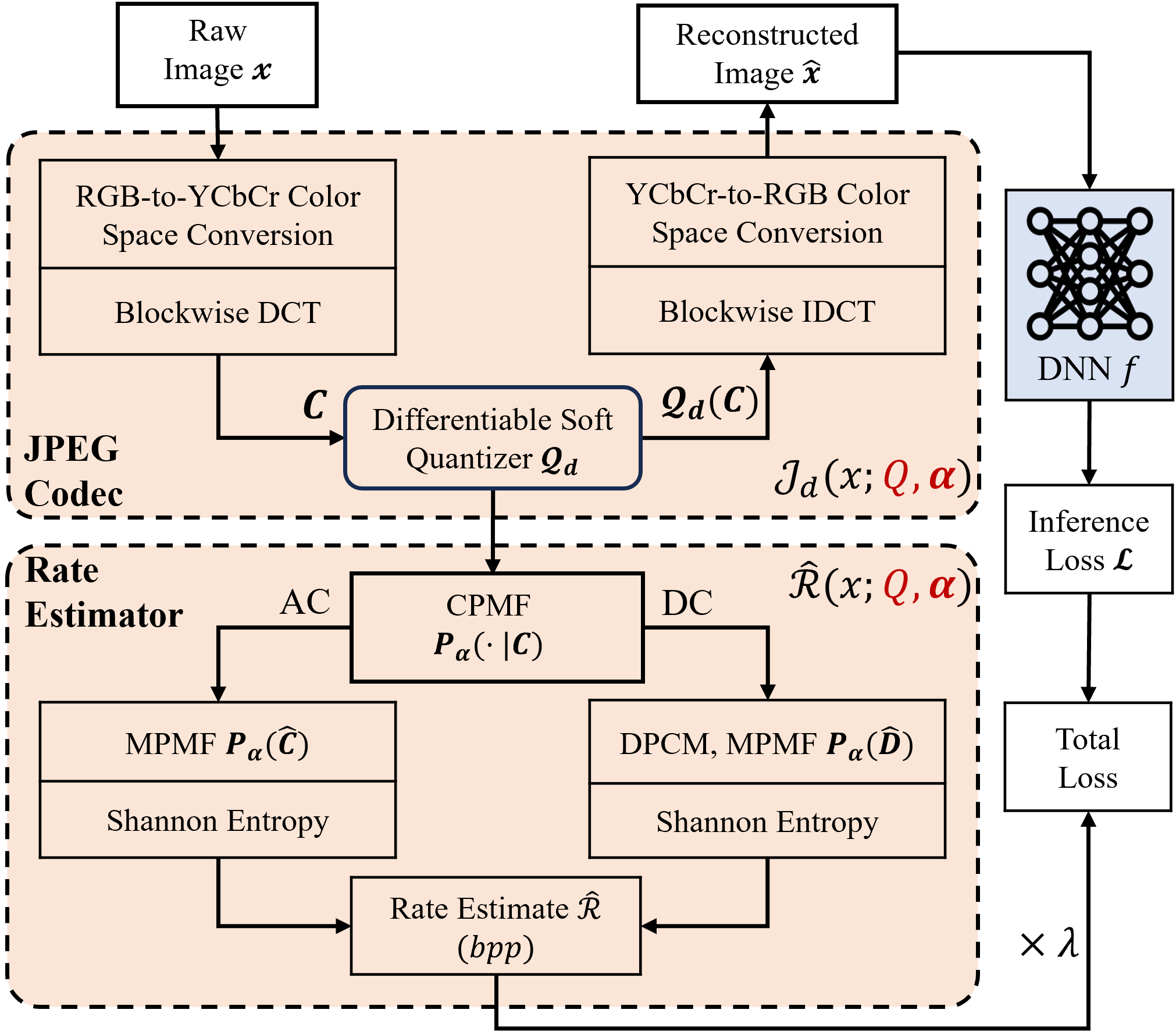}
    \caption{The overall training framework of J4D is illustrated, whose major components are a differentiable JPEG codec and an information-theoretic rate estimator. Note that the quantization tables $Q$ and $\alpha$ tables $\boldsymbol{\alpha}$, shown in red, are the only trainable parameters, whereas the pre-trained DNN $f$ is always frozen.}
    \vspace{-2ex}
    \label{fig:workflow}
\end{figure}

Image compression is a fundamental aspect of image processing and computer vision (CV). For decades, JPEG \cite{wallace1992jpeg} has stood out and become the \textit{de facto} choice among various image compression techniques, due to its efficacy and simplicity in reducing image size while maintaining visual quality for human perception. As a lossy compression technique, JPEG exploits the characteristics of the human visual system to discard information which is typically imperceptible to the human eye.

However, with the rise of deep learning (DL), there has been a significant shift in the way image data is consumed. In the past decade, deep neural networks (DNNs) have become central to numerous CV tasks, including image classification \cite{simonyan2014very, he2016deep, krizhevsky2012imagenet}, semantic segmentation \cite{long2015fully, Ronneberger2015UNetCN, he2017mask}, etc. Consequently, an increasing number of images, potentially compressed by JPEG, are now processed by DNNs with minimal human intervention. Since DNNs perceive images differently from humans, JPEG compression designed for human vision often leads to significant performance degradation for DNN-based applications \cite{liu2018deepn, xie2019source, bhowmik2022lost}. As a result, given the predominance of JPEG in image compression and the ubiquity of DNNs in CV applications, there is a pressing need to develop JPEG-compatible compression that better aligns with DNN vision.

To address the above issue, existing works such as \cite{xie2019source, Zheng2023jpegcompliant, Salamah2024jpegcompliant} promote two-step methods, where one first extracts the sensitivity of DNNs with respect to different Discrete Cosine Transform (DCT) frequencies, and then recasts the conventional rate-distortion optimization problem for human vision into rate-accuracy optimization problems for DNN vision based on the sensitivity, whose solutions are the designed JPEG encoding parameters for DNNs. However, these methods rely on a first-order Taylor approximation of DNN loss functions, which becomes inaccurate when the quantization error is large, leading to unsatisfactory performance at a low compression rate where compressed images are highly distorted. On the other hand, trainable JPEG approaches such as \cite{xie2022bandwidth, Luo2020ratedistortion} approximate the JPEG codec with differentiable proxies and estimate the compression rate using either a small neural network or straight-through estimators (STEs), and then train JPEG by optimizing objectives involving both DNN inference performance and compression rate using backpropagation. However, due to their suboptimal approximation of the JPEG codec and reliance on rough rate estimators, they cannot consistently outperform the default JPEG. In addition, none of the above methods manages to design universal JPEG encoding parameters for various DNN architectures.

To address the limitation of existing methods, this paper introduces a novel trainable JPEG solution, dubbed J4D, based on a probabilistic quantization quantizer, from which both a differentiable JPEG codec and an analytical rate estimate can be derived. By taking the conditional expectation of the probabilistic quantizer, we obtain a differentiable soft quantizer, which substitutes the non-differentiable uniform quantizer in the default JPEG to enable gradient update for JPEG encoding parameters. The compression rate, on the other hand, can be closely estimated by the sum of entropies of quantized DCT coefficients from different frequencies, whose marginal probability mass functions (MPMFs) result directly from the probabilistic quantizers. Given these differentiable JPEG codec and information theory-based rate estimator, we then train JPEG to determine encoding parameters for DNN vision through backpropagation. After training, the probabilistic quantizer equipped with the learned encoding parameters will be subsequently used in actual JPEG encoding. As a result, during the actual JPEG encoding phase, J4D has roughly the same computation complexity as the default JPEG. Extensive experiments on three datasets, covering both image classification and semantic segmentation tasks, and seven DNN architectures validate the effectiveness of J4D, showing consistent and significant improvements over the default JPEG and other JPEG codecs designed in the literature specifically for DNNs. Specifically, on a fine-grained classification dataset, J4D improves the Top-1 accuracy of the default JPEG by up to 11.60\% at the same rate, or reduces the compression rate of the default JPEG by 80.05\% at the same accuracy. Moreover, to the best of our knowledge, we are the first to design universal JPEG encoding parameters for a group of DNNs with distinct architectures, which deliver comparable rate-accuracy (R-A) performances to those achieved by customized parameters for individual DNNs.

The main contributions of this paper are as follows:
\begin{itemize}
    \item Exploiting probabilistic quantization, we develop a novel rate estimator for JPEG compression with a nice analytical formula.
    \item Based on the new rate estimator and a differentiable JPEG codec, we propose a new training framework to optimize JPEG encoding parameters for DNN vision, dubbed J4D.
    \item The outstanding performance of J4D is validated by comprehensive experiments on multiple CV tasks, datasets, and DNN architectures.
    \item It is shown for the first time that it is possible to design, within J4D, universal JPEG encoding parameters for multiple DNN architectures.
\end{itemize}

\section{Related Work} \label{sec:Related_work}

In the literature, there are several lines of research works on JPEG compression for DNN vision. The first line is known as DNN-friendly JPEG compression, as exemplified in \cite{liu2018deepn, li2020optimizing}.  For example, DeepN-JPEG \cite{liu2018deepn} claims that the standard deviation of DCT coefficients at each frequency position determines the importance of this frequency to DNN inference performance. Therefore, DeepN-JPEG uses a heuristic, piece-wise linear function to map those standard deviations to quantization step sizes for corresponding DCT frequencies, resulting in DNN-oriented JPEG quantization tables. Li \etal \cite{li2020optimizing} recommend a sorted random search algorithm over the entire quantization table space to find better quantization tables for DNN vision than those used in the default JPEG. However, these approaches purely rely on heuristic algorithms without solid justifications, and they also fail to leverage any characteristics of target DNN models, thus yielding suboptimal performance.

The second line promotes sensitivity-based methods. GRACE \cite{xie2019source} evaluates DNN perceptual sensitivity by analyzing the gradient magnitude of its training loss function with respect to different DCT frequencies and color channels, and then uses the sensitivity to guide the optimization of JPEG quantization tables and color space conversion. Zheng \etal \cite{Zheng2023jpegcompliant} and Salamah \etal \cite{Salamah2024jpegcompliant}, on the other hand, propose a new distortion measure for DNN vision built upon DNN's sensitivity to perturbations at different DCT frequencies, based on which they recast existing quantization table design algorithms for human vision into those for DNN vision. However, both of the above methods rely heavily on a first-order Taylor approximation of DNN loss functions, which may not be valid in the case of a low compression rate due to large quantization steps.

The third line focuses on training JPEG encoding for DNN vision through backpropagation. To realize this, two fundamental problems have to be addressed properly. The first problem is how to convert the non-differentiable JPEG codec into a differentiable one before it can be prepended to DNNs for gradient update. The second problem, which is more challenging, is how to represent the compression rate of JPEG analytically so that it can be minimized using gradient-based methods as well. The approach described in \cite{Luo2020ratedistortion}, dubbed Google throughout this paper (since it's proposed by Google Research), employs a third-order polynomial approximation of the rounding function to make JPEG differentiable and uses a 3-layer neural network to estimate the compression rate. AutoJPEG \cite{xie2022bandwidth} adopts a tanh-based approximation of the rounding operation to construct a differentiable JPEG codec and builds a rate estimator based on the run-length encoding (RLE) mechanism. To enable gradient propagation during training from the rate estimate back to JPEG encoding parameters, AutoJPEG repeatedly uses STEs and other approximations to bypass non-differentiability.  Although both approaches indeed provide trainable JPEG frameworks, they rely on rough approximations extensively, hence leading to unsatisfactory performance inevitably. Nonetheless, trainable JPEG approaches generally outperform methods in the previous two categories, since they successfully leverage online gradient descent for solving the optimization problem. More related work is discussed in \cref{sec:supp_related_work} in the supplementary material.

Our proposed method falls into the trainable JPEG category but solves the two fundamental problems mentioned above analytically. First, to break through all the constraints imposed by the conventional non-differentiable uniform quantization used in JPEG, we use probabilistic quantization, which in turn leads to differentiable soft quantizers. Second, with the help of probabilistic quantization, we introduce a neat analytical rate estimator. Furthermore, J4D is able to integrate the resulting differentiable JPEG codec and the rate estimator into a cohesive system, instead of piecing ad hoc approximation modules together. Since all components within the pipeline are now analytical, J4D can fully unleash the potential of gradient-based optimization to get the optimal solution for JPEG encoding parameters. Additionally, in actual JPEG encoding, J4D has roughly the same computation complexity as the default JPEG.

It should be noted that differentiable soft quantizers used in J4D have been employed in \cite{Salamah2024jpegdl} recently. However, the problem addressed in this paper is different from that in \cite{Salamah2024jpegdl}. In their case, the differentiable JPEG codec is used to improve DNN generalization performance without considering compression rate at all, whereas we care about the R-A trade-off rather than accuracy alone, and rate estimation is a big issue.

\section{Problem Formation} \label{sec:Problem_Formation}

We denote a raw RGB image as $x \in \mathbb{R}^{3 \times W \times H}$, where $W$ and $H$ correspond to the width and height of the input image. In the JPEG pipeline, $x$ is first converted into the YCbCr color space followed by non-overlapping partitioning into $N$ blocks with size $8 \times 8$ for each channel. Then, DCT is applied to each block to obtain the corresponding DCT coefficients. After flattening the DCT coefficients in zig-zag order, we obtain $M = 64$ sequences of DCT coefficients ordered from low to high frequency. We denote these DCT coefficients as $\mathbf{C}$ = $[C^{l}_{m,n}]$, where $l=1,2,3$ correspond to color channels Y, Cb and Cr respectively, $1 \leq m \leq M$ is the frequency position index, and $1 \leq n \leq N$ is the block index. Note that all operations involved so far are differentiable. The next step is quantization. Let $Q_Y$ = $[q_1, q_2, \dots, q_{M}]$ and $Q_C$ = $[q_{M+1}, q_{M+2}, \dots, q_{2M}]$ denote the quantization tables used for the luminance (Y) and chrominance (CbCr) channels respectively. In the default JPEG, each DCT coefficient $C^{l}_{m, n}$ is quantized by the uniform quantization following $\mathcal{Q}_u(C^{l}_{m, n};q_m)=\lfloor{C^{l}_{m, n}/q_m}\rceil \times q_m$ for $l = 1$ and $\mathcal{Q}_u(C^{l}_{m, n};q_{M+m})=\lfloor{C^{l}_{m, n}/q_{M+m}}\rceil \times q_{M+m}$ for $l = 2,3$, where $\mathcal{Q}_u$ denotes the uniform quantizer. Note that with respect to the quantization step size $q$, uniform quantization is either non-differentiable at each jump point, or has a zero derivative, making it unsuitable for gradient updation. After that, lossless coding including RLE and entropy coding, \eg, Huffman coding, is used to encode the quantized DCT coefficients into a bit stream, whose complexity is often measured by the compression rate $\mathcal{R}(x; Q)$ in bit-per-pixel (bpp), where $Q=(Q_Y,Q_C)$. In the reconstruction stage, the bit stream is decoded back to the quantized DCT coefficients, after which inverse operations including blockwise inverse DCT (IDCT) and YCbCr-to-RGB color space conversion are conducted to get the reconstructed RGB image $\hat{x}$. For simplicity, we denote the entire JPEG codec as $\mathcal{J}_u$, where the subscript indicates the application of the uniform quantizer $\mathcal{Q}_u$, hence $\hat{x}=\mathcal{J}_u(x; Q)$.

Given a pre-trained DNN model $f$, we aim to design quantization tables to maximize its inference accuracy on $\hat{x}=\mathcal{J}_u(x; Q)$ subject to a constraint over the compression rate. This can be achieved by minimizing the Lagrangian composed of the training loss function of $f$ and the compression rate, formalized as 
\begin{small}\begin{align} \label{eq:Total_loss}
    \min_{Q} \mathbb{E}	\left[ \mathcal{L} ( f(\mathcal{J}_u(x;Q)), y) + \lambda \mathcal{R}(x;Q) \right],
\end{align}\end{small}where $\mathcal{L}$ denotes the loss function used to train $f$, $y$ is the ground truth label of $x$, and $\lambda$ is a tunable hyper-parameter that controls the trade-off between compression rate and DNN performance.

However, to solve the above optimization problem with gradient descent, two key challenges are (1) the non-differentiability within $\mathcal{J}_u$ due to uniform quantization, and (2) the need for an analytical representation of $\mathcal{R}(x;Q)$, the solutions to which will be discussed in the next section.

\section{Methodology} \label{sec:Methodology}

In the first part of this section, we first invoke two types of quantizers from \cite{Yang2024systems} and then discuss how to incorporate them into the JPEG pipeline to construct a differentiable JPEG codec. In the second part of this section, we present our information-theoretic rate estimator $\mathcal{\hat{R}}(\cdot)$ and then summarize the overall J4D framework.

\subsection{Probabilistic and Differentiable Soft Quantizer} \label{sec:Qp_Qd}
Denote a symmetric index set of uniform quantization with length $2L+1$ as:
\begin{small}\begin{align} \label{eq:A}
    \mathcal{A} = \{ -L, -L+1, \dots, 0, \dots, L-1, L \}.
\end{align}\end{small}For convenience, $\mathcal{A}$ is also regarded as a vector. Multiplying $\mathcal{A}$ by a quantization step size $q$, we get the corresponding reconstruction space:
\begin{small}\begin{align}\label{eq:recon_space}
    \hat{\mathcal{A}} = [ -L, -L+1, \dots, 0, \dots, L-1, L ] \times q.
\end{align}\end{small}We again regard $\hat{\mathcal{A}}$ as both a vector and a set.

Let $C$ represent a DCT coefficient. To quantize $C$ randomly to some $\hat{C}$ = $iq \in \hat{\mathcal{A}}$ with $i \in \mathcal{A}$, we invoke from \cite{Yang2024systems} a probabilistic quantizer $\mathcal{Q}_p$. As described in \cite{Yang2024systems}, a conditional probability mass function (CPMF) over the reconstruction space $\hat{\mathcal{A}}$ or, equivalently, the index set $\mathcal{A}$ given $C$ is defined as
\begin{small}\begin{align}  \label{eq:CPMF_1}
    P_{\alpha}(\hat{C}|C) = \frac{e^{-\alpha (C - \hat{C})^2}}{\sum_{j \in \mathcal{A}} e^{-\alpha (C - jq)^2 }}, \forall \hat{C} \in \mathcal{\hat{A}},
\end{align}\end{small}where $\alpha > 0$ is a trainable scaling factor.

Define $\big[C \big]_{2L+1} = [\overbrace{C, \dots, C}^{\text{$2L+1$ times}}]$. Then, the CPMF $P_{\alpha}(\cdot|C)$, as a probability vector with dimension $2L+1$, can be easily computed using the softmax operation $\boldsymbol{\sigma}(\cdot)$:
\begin{small}\begin{align}   \label{eq:CPMF_2}
    \big[P_{\alpha}(\cdot | C)\big] = \boldsymbol{\sigma} \left( - \alpha \left( \big[C\big]_{2L+1} -\hat{\mathcal{A}} \right)^2 \right).
\end{align}\end{small}

With the CPMF, $C$ is now quantized to each $\hat{C} \in \hat{\mathcal{A}}$ with probability $P_{\alpha}(\hat{C}|C)$, and this random mapping is referred to as the probabilistic quantizer $\mathcal{Q}_p$.
Note that given $C$, $\mathcal{Q}_p(C;q,\alpha)$ is a random variable with distribution $P_{\alpha}(\cdot | C)$ over the alphabet $\hat{\mathcal{A}}$.

Due to the random jump, $\mathcal{Q}_p$ is non-differentiable. However, taking the conditional expectation of $\mathcal{Q}_p(C)$ given $C$, we then get a differentiable soft quantizer 
\begin{small}\begin{align} \label{eq:Qd}
    \mathcal{Q}_d(C) = \mathbb{E} [\mathcal{Q}_p(C)|C] = \sum_{\hat{C} \in \hat{\mathcal{A}}} \hat{C} \cdot P_{\alpha}(\hat{C}|C).
\end{align}\end{small}As $\alpha \rightarrow \infty$, $\mathcal{Q}_d$ approaches $\mathcal{Q}_u$, as shown in \cref{fig:Q_u_Q_d} in supplementary materials.

\subsection{Differentiable JPEG Codec} \label{sec:diff_JPEG}
As shown in \cref{eq:recon_space}, given a predetermined index set, the reconstruction space solely depends on the quantization step size. In JPEG, there are in total $2M$ quantization steps, so the corresponding reconstruction spaces are
\begin{small}\begin{align}
    \hat{\mathcal{A}}_k &= [-L,-L+1,\dots,0,\dots,L-1,L] \times q_k,
\end{align}\end{small}where $1 \leq k \leq 2M$. Then, we can incorporate $2M$ probabilistic quantizers, \ie, $\mathcal{Q}_p(C^{l}_{m,n}; q_{m}, \alpha_{m}) \in \hat{\mathcal{A}}_m$ for $l = 1$ and $\mathcal{Q}_p(C^{l}_{m,n}; q_{M+m}, \alpha_{M+m}) \in \hat{\mathcal{A}}_{M+m}$ for $l = 2,3$, into JPEG to replace uniform quantizers, thereby creating a probabilistic JPEG codec $\mathcal{J}_p(x; Q, \boldsymbol{\alpha})$, where $\boldsymbol{\alpha} = (\boldsymbol{\alpha}_Y, \boldsymbol{\alpha}_C)$, $\boldsymbol{\alpha}_Y = [\alpha_1, \alpha_2, \dots, \alpha_M]$ and $\boldsymbol{\alpha}_C = [\alpha_{M+1}, \alpha_{M+2}, \dots, \alpha_{2M}]$ are $\alpha$ tables for the Y and CbCr channels, respectively.

Further applying \cref{eq:Qd}, we get $2M$ differentiable soft quantizers accordingly, \ie, $\mathcal{Q}_d(C^{l}_{m,n}; q_{m}, \alpha_{m})$ for $l = 1$ and $\mathcal{Q}_d(C^{l}_{m,n}; q_{M+m}, \alpha_{M+m})$ for $l = 2,3$. The resulting JPEG codec $\mathcal{J}_d(x; Q, \boldsymbol{\alpha})$, where $\mathcal{Q}_u$'s in the default JPEG are substituted by $\mathcal{Q}_d$'s, is differentiable everywhere, effectively enabling gradient propagation.

\subsection{Entropy-based Rate Estimation} \label{sec:rate}
Thanks to $\mathcal{Q}_p$, we can provide an estimate $\mathcal{\hat{R}}$ of the actual compression rate $\mathcal{R}$ based on the Shannon entropy of the randomly quantized DCT coefficients, as detailed below.

Given any DCT coefficient $C^{l}_{m,n}$ and the $\mathcal{Q}_p$ associated with it, we immediately get the CPMF of the quantized DCT coefficient following \cref{eq:CPMF_2}. For any $m$, note that $\mathcal{Q}_p( C^{1}_{m,n};q_m,\alpha_m)$ ($\mathcal{Q}_p( C^{l}_{m,n};q_{M+m},\alpha_{M+m})$, $l=2, 3$, resp.), $1 \leq n \leq N$, are independent random variables. It can be shown that the expectations of the empirical distributions of randomly quantized  DCT coefficients in frequency $m$ of Y or CbCr channel are given by the marginal probability mass functions (MPMFs) of the quantized DCT coefficients:
\begin{small}\begin{align} \label{eq:MPMF}
    P_{\alpha_m}(\hat{C}) 
    &= \frac{1}{N} \sum^{N}_{n=1} P_{\alpha_{m}}(\hat{C}| C^{1}_{m,n}), \forall \hat{C} \in \hat{\mathcal{A}}_m, \\
    P_{\alpha_{M+m}}(\hat{C}) &= \frac{1}{2N} \sum^{3}_{l = 2} \sum^{N}_{n = 1} P_{\alpha_{M+m}}(\hat{C} | C^{l}_{m,n}), \forall \hat{C} \in \hat{\mathcal{A}}_{M+m}, 
\end{align}\end{small}where $2 \leq m \leq M$. Note that two DC frequencies are left out for special treatment.

When entropy coding, \eg, Huffman coding, is adopted to encode the quantized DCT coefficients, the average codeword length (in bits) per coefficient can be closely approximated by the Shannon entropy $\mathcal{H}(\cdot)$ of the probability distributions of the quantized DCT coefficients. As a result, for AC frequencies, we have our rate estimates (in bits):
\begin{small}\begin{align} \label{eq:R_ac}
    \hat{\mathcal{R}}_{m} = N \cdot \mathcal{H}(P_{\alpha_{m}}(\cdot)), \hat{\mathcal{R}}_{M+m} = 2N \cdot \mathcal{H}(P_{\alpha_{M+m}}(\cdot)),
\end{align}\end{small}
where $2 \leq m \leq M$.

For DC frequencies, differential pulse code modulation (DPCM) is applied in the JPEG pipeline to encode the differences between the quantized DC coefficients of adjacent blocks rather than the quantized coefficients themselves. This leads to a distinct rate estimation for the DC frequencies, as discussed below.

Given two adjacent DC coefficients, we first compute the respective CPMFs of the differentially encoded quantized DC coefficients by applying \cref{eq:CPMF_2} twice. Then, by convolution, the CPMFs of differentially encoded quantized DC coefficients $\hat{D}$ are given by
\begin{small}\begin{align}
    &P_{\alpha_1}(\hat{D} | C^{1}_{1,n-1}, C^{1}_{1,n}) \nonumber \\ 
    &= \sum_{\hat{C} \in \hat{\mathcal{A}}_1} P_{\alpha_1}(\hat{C} | C^{1}_{1,n}) \cdot P_{\alpha_1}(\hat{C}-\hat{D} | C^{1}_{1,n-1}), \nonumber \\
    &\forall \hat{D} \in \{a-b | a, b \in \hat{\mathcal{A}}_1 \},
\end{align}\end{small}for the luminance channel, and
\begin{small}\begin{align}
    &P_{\alpha_{M+1}}(\hat{D} | C^{l}_{1,n-1}, C^{l}_{1,n}) \nonumber \\ 
    &= \sum_{\hat{C} \in \hat{\mathcal{A}}_{M+1}} P_{\alpha_{M+1}}(\hat{C} | C^{l}_{1,n}) \cdot P_{\alpha_{M+1}}(\hat{C}-\hat{D} | C^{l}_{1,n-1}), \nonumber \\
    &\forall \hat{D} \in \{a-b | a, b \in \hat{\mathcal{A}}_{M+1} \}, l=2,3,
\end{align}\end{small}for chrominance channels, where $1 \leq n \leq N$, and note that $C^l_{1,0}:=0$ for all $l$. Once again, it can be shown that the expectations of the empirical distributions of differentially encoded quantized DC coefficients are given by the following MPMFs:
\begin{small}\begin{align}
    P_{\alpha_1}(\cdot) &= \frac{1}{N} \sum^{N}_{n=1} P_{\alpha_1}(\cdot | C^{1}_{1,n-1}, C^{1}_{1,n}), \\
    P_{\alpha_{M+1}}(\cdot) &= \frac{1}{2N} \sum^{3}_{l=2} \sum^{N}_{n=1} P_{\alpha_{M+1}}(\cdot | C^{l}_{1,n-1}, C^{l}_{1,n}). 
\end{align}\end{small}Similar to \cref{eq:R_ac}, we obtain rate estimates for DC frequencies as
\begin{small}\begin{align} \label{eq:R_dc}
    \hat{\mathcal{R}}_{1} = N \cdot \mathcal{H}( P_{\alpha_1}(\cdot)),
    \hat{\mathcal{R}}_{M+1} = 2N \cdot \mathcal{H}( P_{\alpha_{M+1}}(\cdot)).
\end{align}\end{small}

Finally, combining \cref{eq:R_ac,eq:R_dc}, the rate estimate of a whole image (in bpp) can be computed by
\begin{small}\begin{align} \label{eq:totbit}
    \mathcal{\hat{R}}(x;Q,\boldsymbol{\alpha}) = \frac{1}{WH} \sum_{k=1}^{2M} \hat{\mathcal{R}}_{k}.
\end{align}\end{small}
\subsection{Overall Framework of J4D} \label{sec:Framework}
Equipped with the differentiable JPEG codec $\mathcal{J}_d$ and the entropy-based rate estimate $\mathcal{\hat{R}}$, we can now rewrite the optimization problem \cref{eq:Total_loss} into
\begin{small}\begin{align}\label{eq:final_Total_loss}
    \min_{Q,\boldsymbol{\alpha}} \mathbb{E}	\left[ \mathcal{L} (f(\mathcal{J}_d(x;Q,\boldsymbol{\alpha})), y) + \lambda \mathcal{\hat{R}}(x;Q,\boldsymbol{\alpha}) \right],
\end{align}\end{small}where the expectation is approximated by the empirical mean over a mini-batch during training. This resulting minimization problem can be solved by gradient descent based on backpropagation since all gradient calculations can be done analytically.

Note that $\mathcal{J}_d$ is only used during training to ensure differentiability, while we switch to $\mathcal{J}_p$ during the validation stage (parameterized by the pre-trained quantization and $\alpha$ tables) to randomly quantize DCT coefficients to better leverage the power of entropy coding.

To summarize this section, we refer readers to \cref{fig:workflow} for an illustration of the overall training framework of J4D.
\section{Experiments} \label{sec:Experiments}

\subsection{Experimental Settings} \label{sec:Exp_setting}
\textbf{Benchmarks}. We compare J4D with the default JPEG and two state-of-the-art (SOTA) trainable JPEG frameworks: Google \cite{Luo2020ratedistortion} and AutoJPEG \cite{xie2022bandwidth}, as discussed in \cref{sec:Related_work}. Since neither framework has publicly released code, we implement Google's framework based on the rate estimator described in \cite{begaint2020compressai} and independently reimplement AutoJPEG. We adhere to the hyperparameter values specified in their respective papers to cover the same rate ranges addressed therein, but we perform additional tuning to extend their coverage to rate ranges not addressed in the original works. Notably, in the original AutoJPEG paper, three quantization tables are jointly trained with color space conversion parameters while DNN weights are fine-tuned. To ensure a fair comparison and maintain JPEG compliance, we optimize only the luminance and chrominance quantization tables within the AutoJPEG framework while keeping other components frozen. Additionally, we include a sensitivity-based method, OptS \cite{Zheng2023jpegcompliant, Salamah2024jpegcompliant}, for comparison in the supplementary material (\cref{sec:sup_opts}).

\textbf{Datasets and DNN models}. We evaluate the R-A performance of two computer vision tasks: image classification and semantic segmentation. For image classification, we adopt two datasets: a fine-grained classification dataset, CUB-200-2011 (CUB200) \cite{wah2011caltech}, and a large-scale dataset, ImageNet-1K \cite{krizhevsky2009learning}. For semantic segmentation, we adopt the Pascal VOC 2012 dataset. For image classification, the tested DNN architectures include ResNet18 \citep{he2016deep}, MobileNetV2 \cite{sandler2018mobilenetv2}, MNasNet \cite{tan2019mnasnet}, ConvNeXt-tiny \cite{Liu2022AConvNet}, and Swin-T \cite{Liu2021Swin}, covering both convolution-based and transformer-based models; for semantic segmentation, we use DeepLabv3-MobileNet and DeepLabv3-ResNet50 \cite{chen2017rethinking}. For CUB200, the pre-trained models are obtained following the training setup outlined in \cite{yun2020regularizing}; for ImageNet-1K, we adopt the pre-trained models from PyTorch \cite{paszke2019pytorch}; for Pascal VOC 2012, the pre-trained models are obtained from \cite{DeepLabV3-repo}.

\textbf{Initialization of Encoding Parameters}.
A key feature of $\mathcal{Q}_d$ is its variable softness based on the trainable $\alpha$. However, the gradient behavior of $Q_d$ with respect to $\alpha$ reveals that when $\alpha$ is sufficiently large, the gradient magnitude for $\alpha$ becomes negligible, as presented in \cite{Salamah2024jpegdl}. This implies that $\boldsymbol{\alpha}$ will not be effectively updated during training if initialized with sufficiently large values. Nonetheless, we do need $\alpha$ to be sufficiently large, since if $\alpha$ decreases, both the quantization error, thus the inference loss, and the compression rate will increase during validation, resulting in a strictly worse performance in the R-A sense. Consequently, in all our experiments, we directly initialize all entries of $\boldsymbol{\alpha}$ to a sufficiently large constant, \ie, 100, and exclude $\boldsymbol{\alpha}$ from the training process (see \cref{sec:alpha_selection} in the supplementary material for more discussion about $\boldsymbol{\alpha}$ initialization).

As for quantization tables, it's observed that the R-A performance of J4D is sensitive to their initialization. Therefore, instead of sticking to a fixed initialization, we initialize quantization tables differently for different rate ranges. Concretely, scaled versions of the default JPEG quantization tables are used as initialization in the low-rate region, corresponding to quality factors (QFs) in the range [5, 90]. For the high-rate region, we leverage the pre-trained model’s sensitivity as derived in \cite{Zheng2023jpegcompliant,Salamah2024jpegcompliant} to initialize quantization tables, thereby favoring a smaller quantization step for a more sensitive frequency to limit the distortion amount on it. Specifically, each quantization table entry is initialized with the reciprocal of its sensitivity and then scaled by a factor $\beta$ within [5, 50] to control the R-A trade-off (see \cref{sec:sup_sens} in the supplementary material for more details). As discussed in \cref{sec:Related_work}, the sensitivity measure is valid only for small distortions, limiting its application to the high-rate region. 

\textbf{Training Settings}. We use the Adam optimizer \cite{Kingma15Adam} with a learning rate of $1 \times 10^{-2}$ for our training. The training setup varies by dataset: for CUB200, quantization tables are trained for 100 epochs on a single GPU with a batch size of 32; for ImageNet-1K, they are trained for 20 epochs using 4 GPUs with a batch size of 64 on each GPU (\ie, 256 in total); for Pascal VOC 2012, training runs for 10000 iterations on a single GPU with a batch size of 16. We vary the R-A trade-off for J4D for each distinct quantization table initialization by setting different $\lambda$ in \cref{eq:final_Total_loss}. Particularly, $\lambda \in [0.1, 50]$ for the low-rate region, where we use QF-based quantization table initialization, and $\lambda \in [0, 5]$ for the high-rate region, where sensitivity-based initialization is adopted. Our final R-A performance, which will be shown in the next subsection, reflects the Pareto frontier of all results we get from different pairs of initial quantization tables and $\lambda$. Note that we clip every quantization step to the range [1, 255] before any forward pass and set $L = 1023$ following the conventions adopted in the default JPEG. Moreover, to reduce the additional training complexity introduced by $\mathcal{Q}_d$, we constrain the support of every CPMF to be the Top-5 closest points in $\mathcal{\hat{A}}$ to the coefficient $C$ being quantized (see \cref{sec:sup_comp} in the supplementary material for more details).

\textbf{Compression Settings}.
For all experiments, input images are preprocessed by resizing and center cropping before being fed into DNNs. For CUB200 and ImageNet-1K, the size of preprocessed images is 224 $\times$ 224, while for Pascal VOC 2012, the size is 513 $\times$ 513. Notably, for J4D and all compared methods, compression is applied to these preprocessed images, treating them as raw inputs. We employ the 4:4:4 chroma subsampling mode, which means there's no chroma subsampling\footnote{We stick to no chroma subsampling since it's shown in \cite{xie2019source} that chroma subsampling typically hurts R-A performance.}. For each dataset, the reported compression rate is computed by averaging the image-wise bpp over the entire validation set. Also, note that we use customized Huffman tables in Huffman coding. Our source code is provided in the supplementary material.

\subsection{Experimental Results}
\begin{figure*}[!ht]
\centering
\subfloat[ResNet18]{\includegraphics[height=0.2\textwidth]{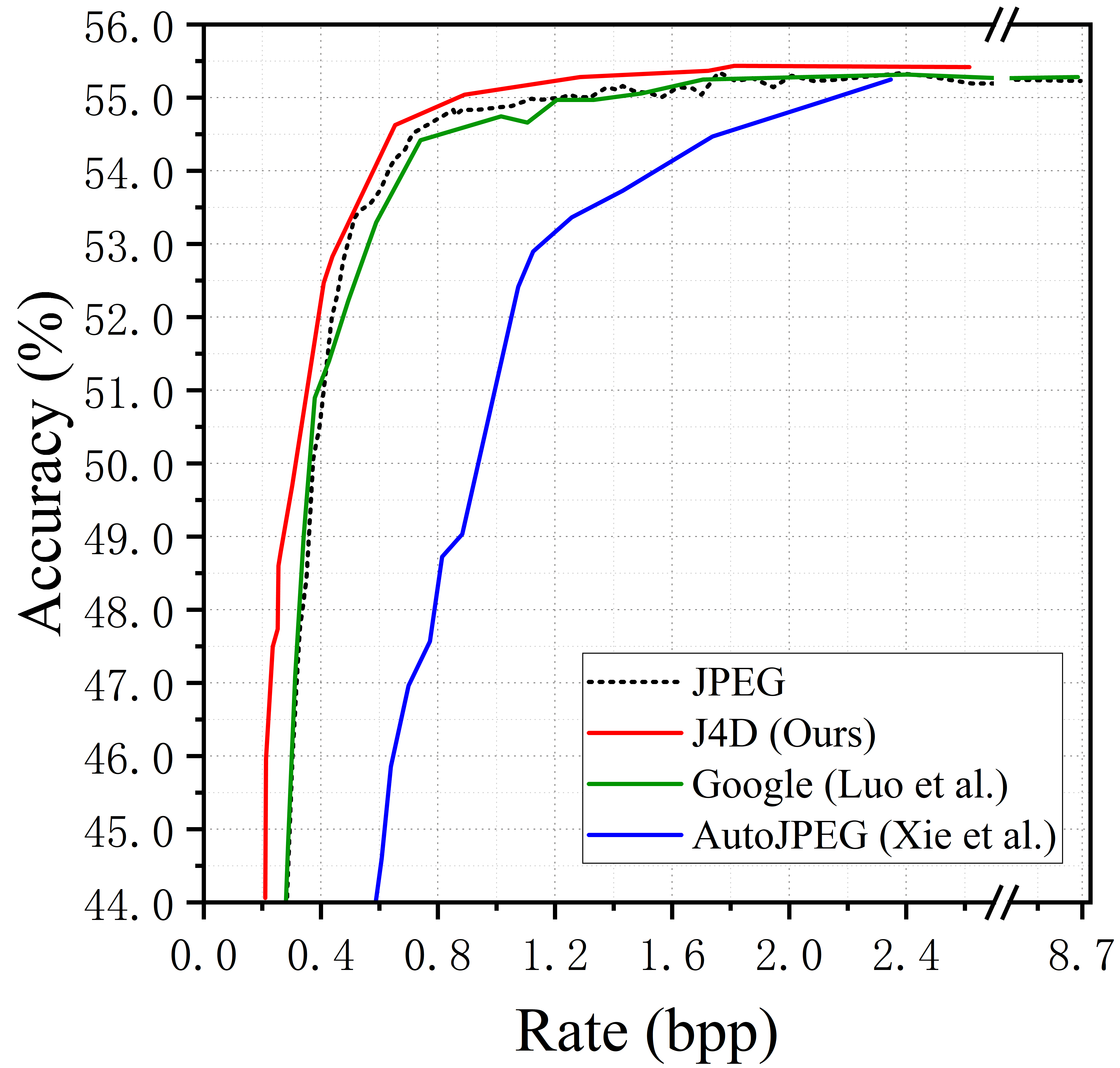}%
\label{fig:cub200_resnet18}}
\hfil
\subfloat[MobileNetV2]{\includegraphics[height=0.2\textwidth]{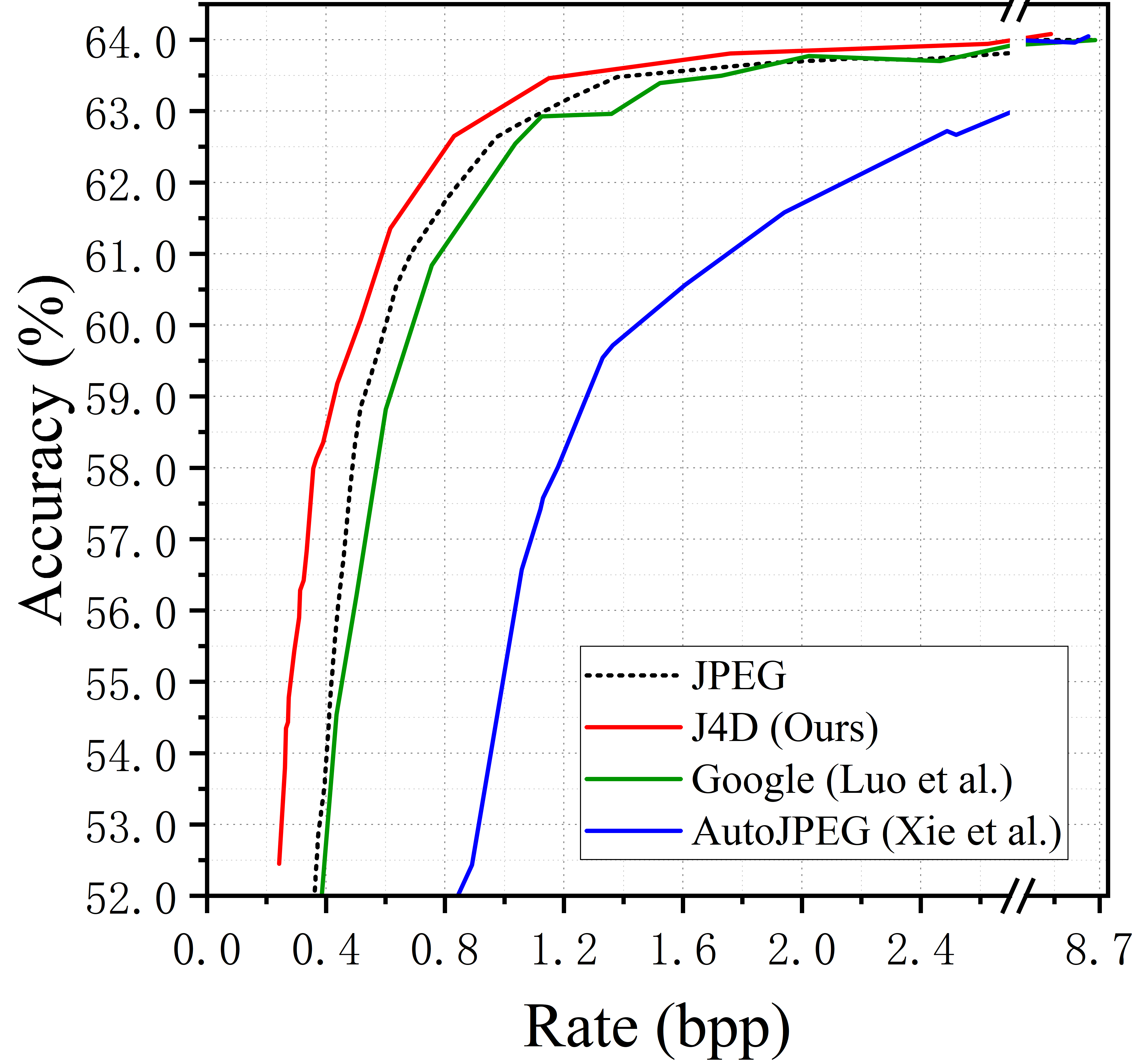}%
\label{fig:cub200_mobilenet_v2}}
\hfil
\subfloat[MnasNet]{\includegraphics[height=0.2\textwidth]{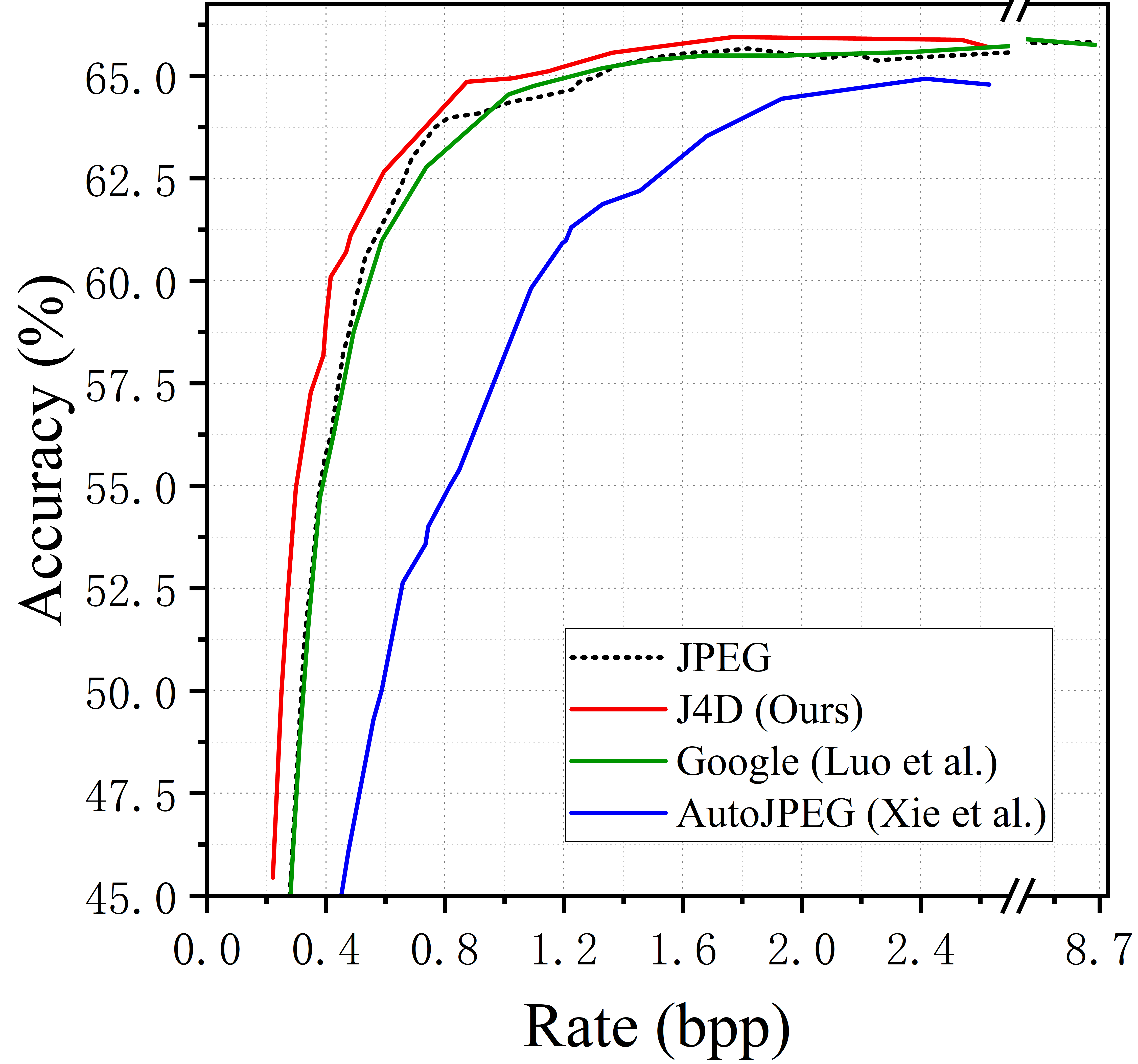}%
\label{fig:cub200_MnasNet}}
\caption{R-A performance on CUB200 for three pre-trained image classification models with architectures (a) ResNet18, (b) MobileNetV2, and (c) MnasNet. Note that the R-A curves are truncated on the right at a rate at which J4D already achieves almost the same level of accuracy as the default accuracy of each model on the original uncompressed dataset.}
\label{fig:CUB200}
\end{figure*}

\textbf{Results on CUB200}. \cref{fig:CUB200} compares the R-A performance of various compression methods across three pre-trained DNN models. The results demonstrate that J4D consistently outperforms all benchmarks across all tested models and over the entire rate range. Compared to the default JPEG, J4D achieves up to an 80.05\% reduction in compression rate without accuracy loss or up to an 11.60\% increase in Top-1 accuracy at a fixed rate. 

\begin{figure*}[!ht]
\centering
\subfloat[MobileNetV2]{\includegraphics[height=0.2\textwidth]{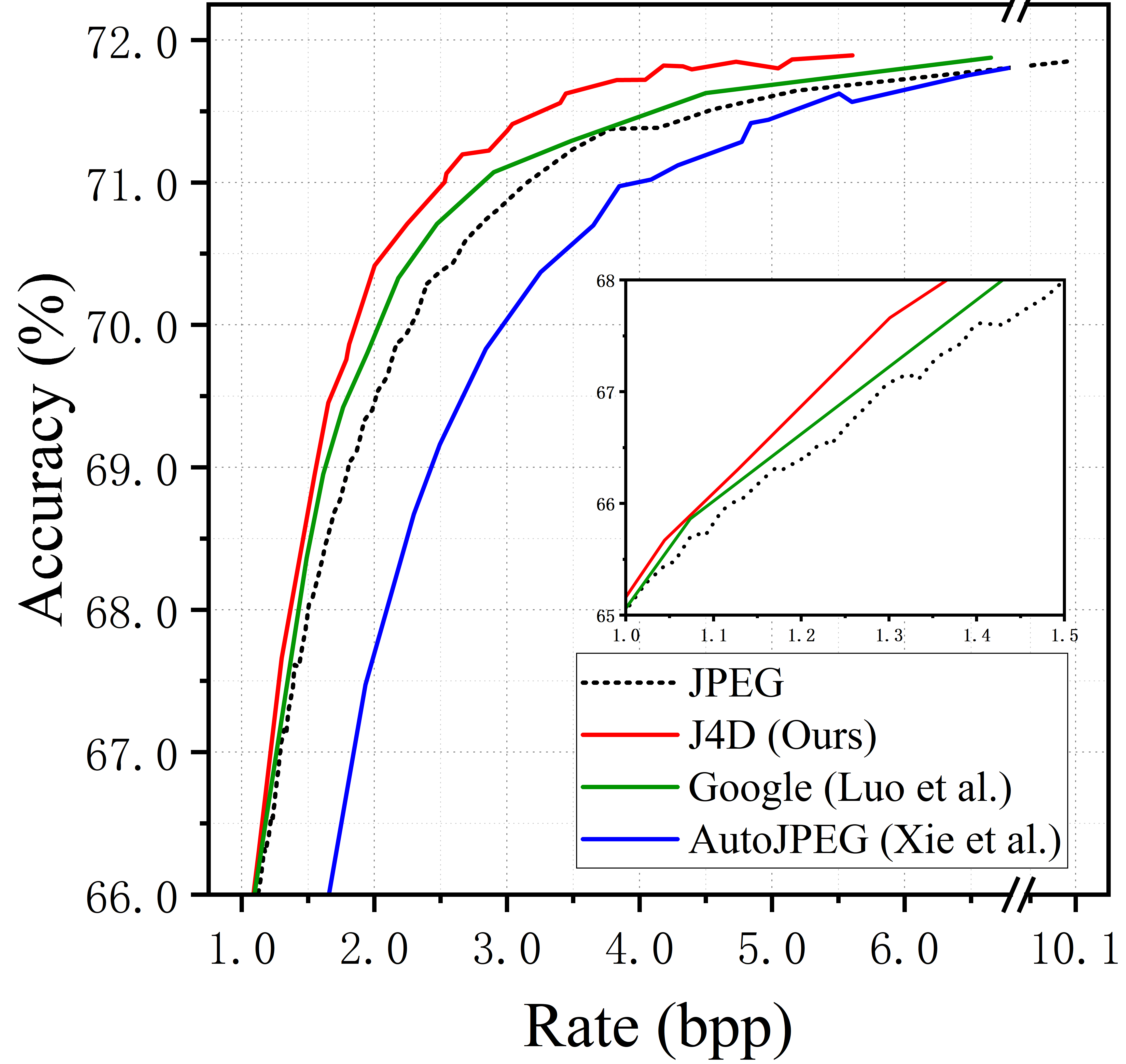}%
\label{fig:imagenet_MobileNetV2}}
\hfil
\subfloat[MnasNet]{\includegraphics[height=0.2\textwidth]{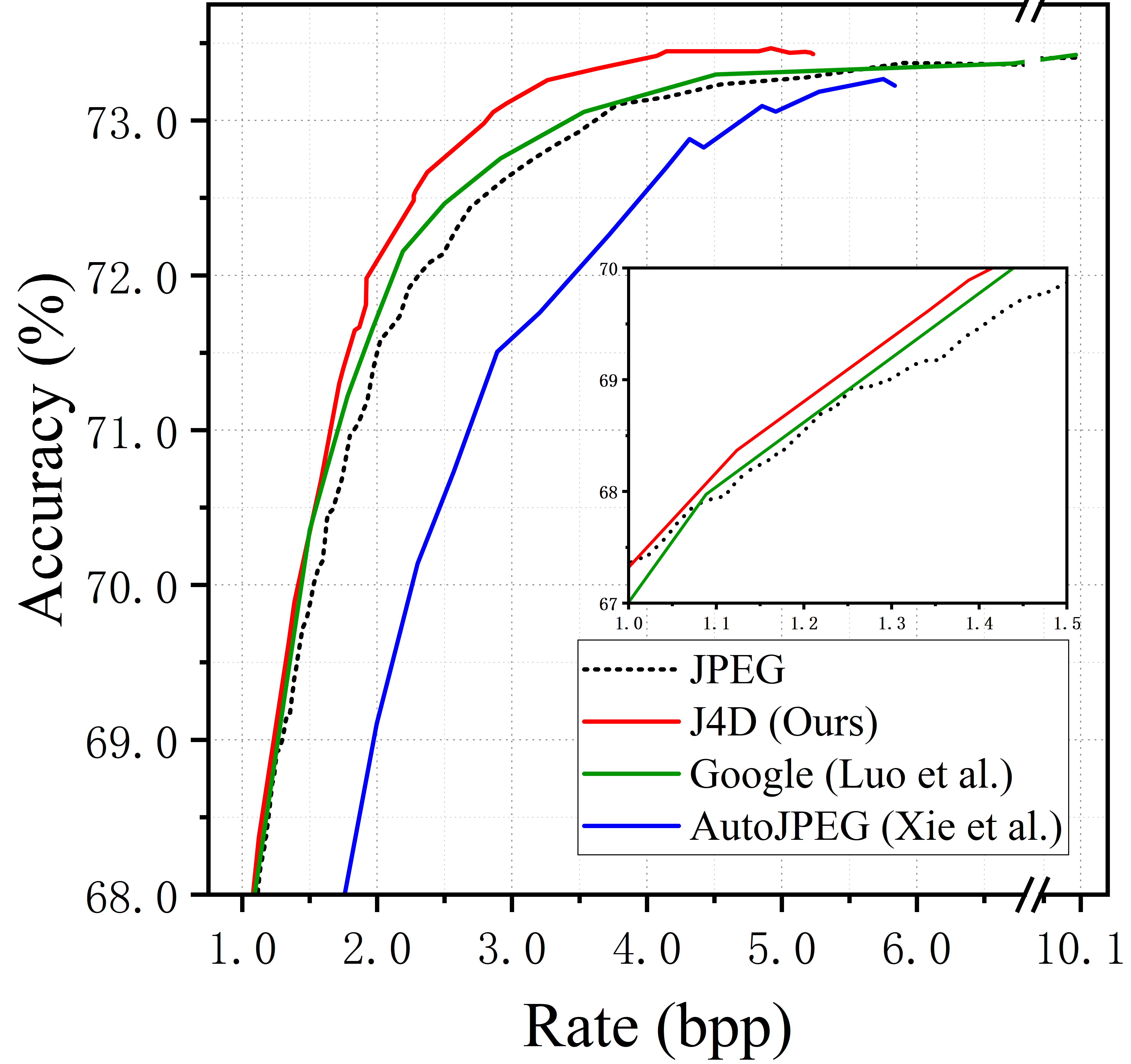}%
\label{fig:imagenet_MnasNet}}
\hfil
\subfloat[ConvNeXt-tiny]{\includegraphics[height=0.2\textwidth]{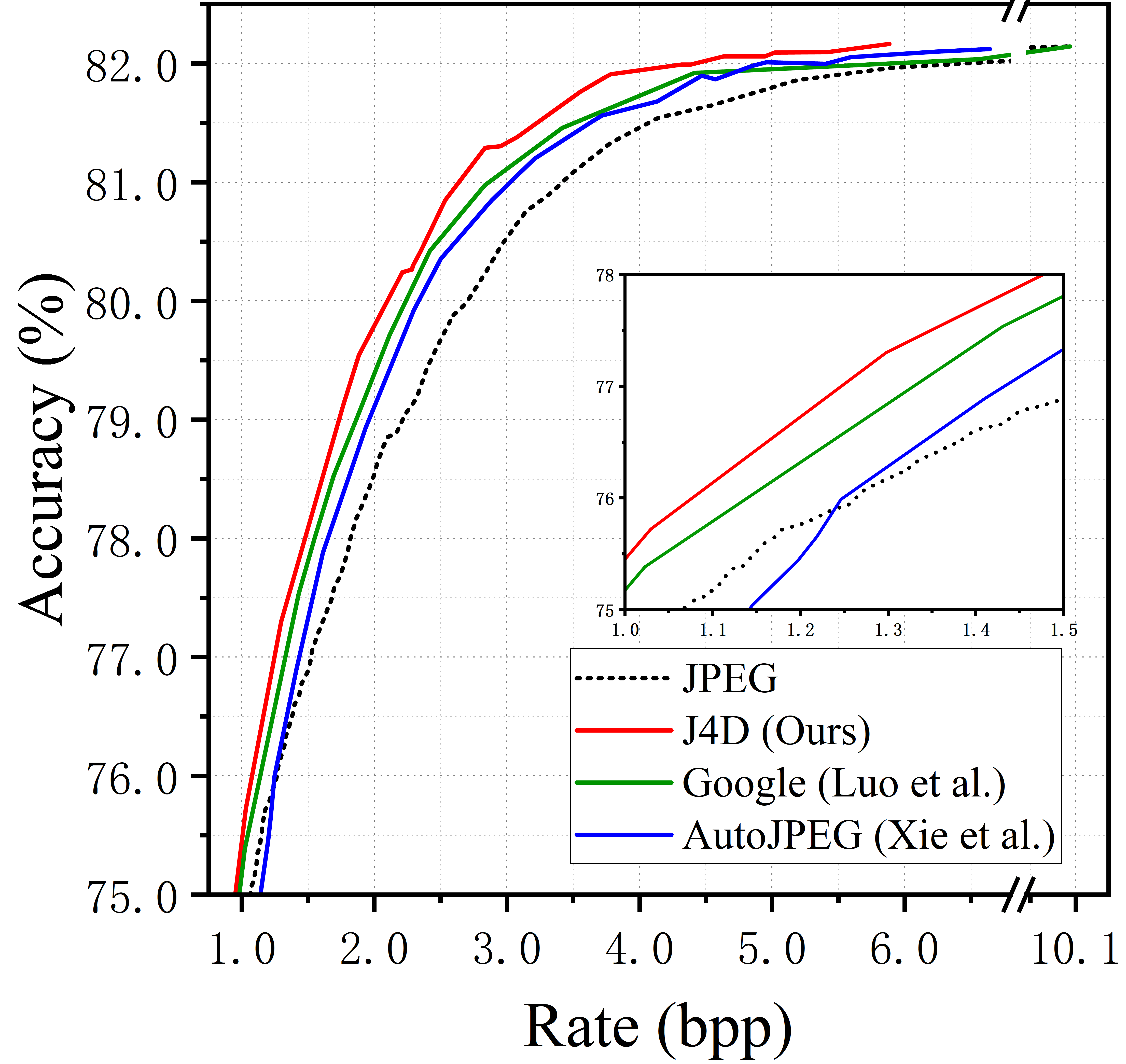}%
\label{fig:imagenet_ConvNeXt_tiny}}
\hfil
\subfloat[Swin-T]{\includegraphics[height=0.2\textwidth]{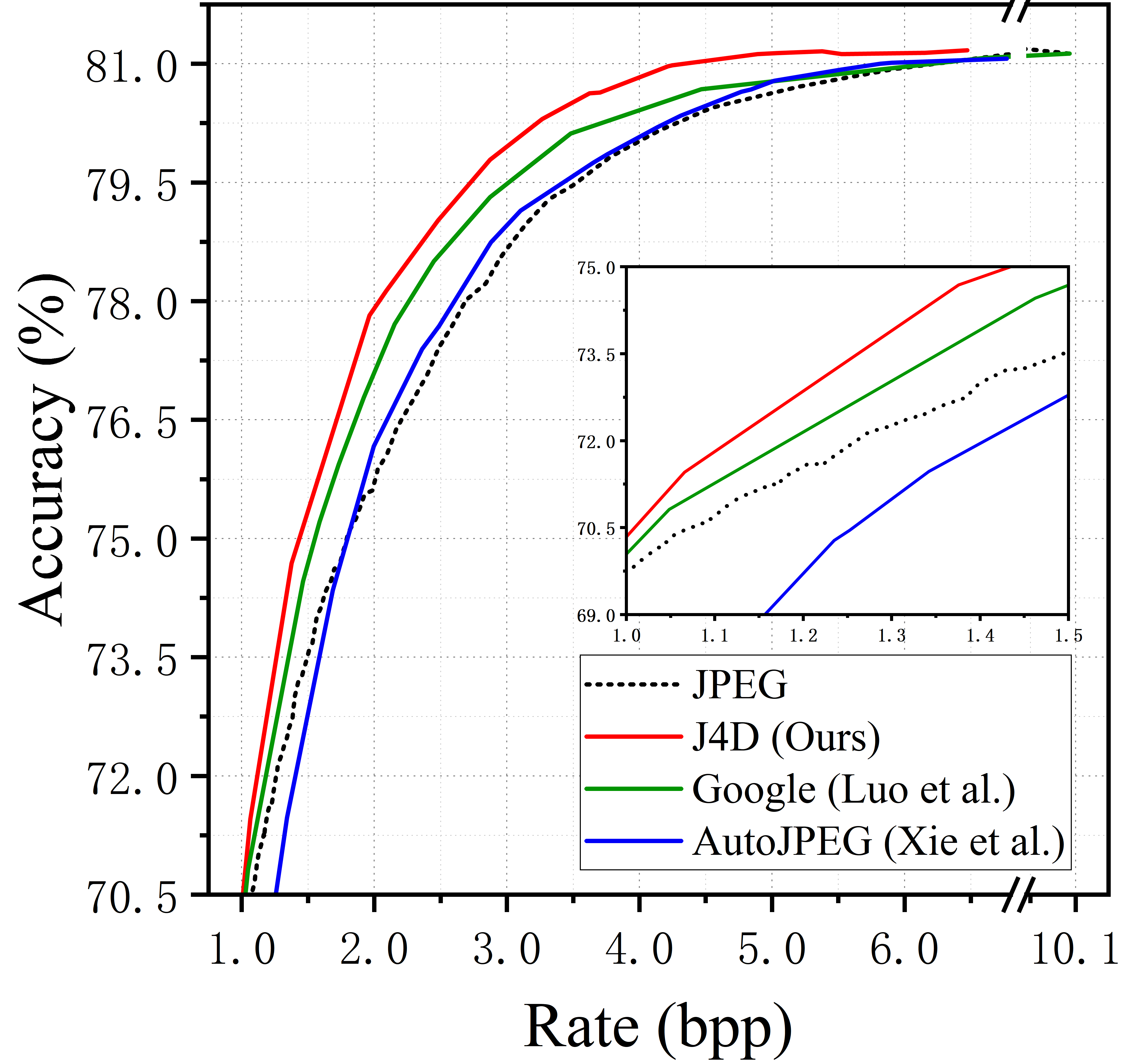}%
\label{fig:imagenet_SwinT}}
\caption{R-A performance on ImageNet-1K for four pre-trained image classification models with architectures (a) MobileNetV2, (b) MnasNet, (c) ConvNeXt-tiny, and (d) Swin-T.}
\label{fig:ImageNet}
\end{figure*}

\textbf{Results on ImageNet-1K}. \cref{fig:ImageNet} demonstrates the consistent superiority of J4D in R-A performance across three convolution-based models and one transformer-based model. Compared to the default JPEG, J4D achieves up to a 59.71\% reduction in rate without accuracy loss or up to a 2.21\% increase in Top-1 accuracy at a fixed rate. 

\begin{figure}[!ht]
    \centering
    \subfloat[DeepLabv3-MobileNet]{\includegraphics[height=0.4\linewidth]{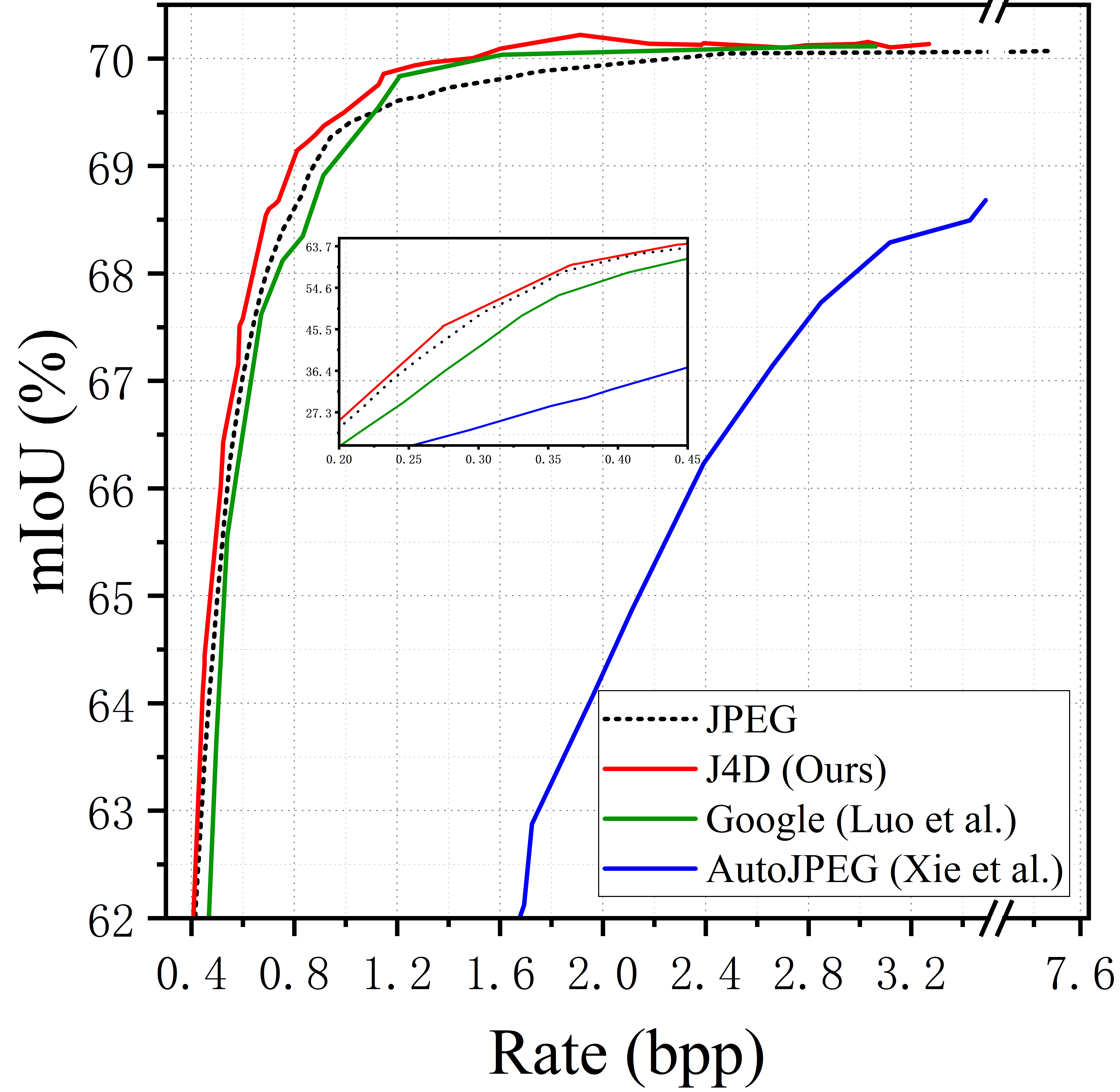}
    \label{fig:VOC_MN}}
    \subfloat[DeepLabv3-ResNet50]{\includegraphics[height=0.4\linewidth]{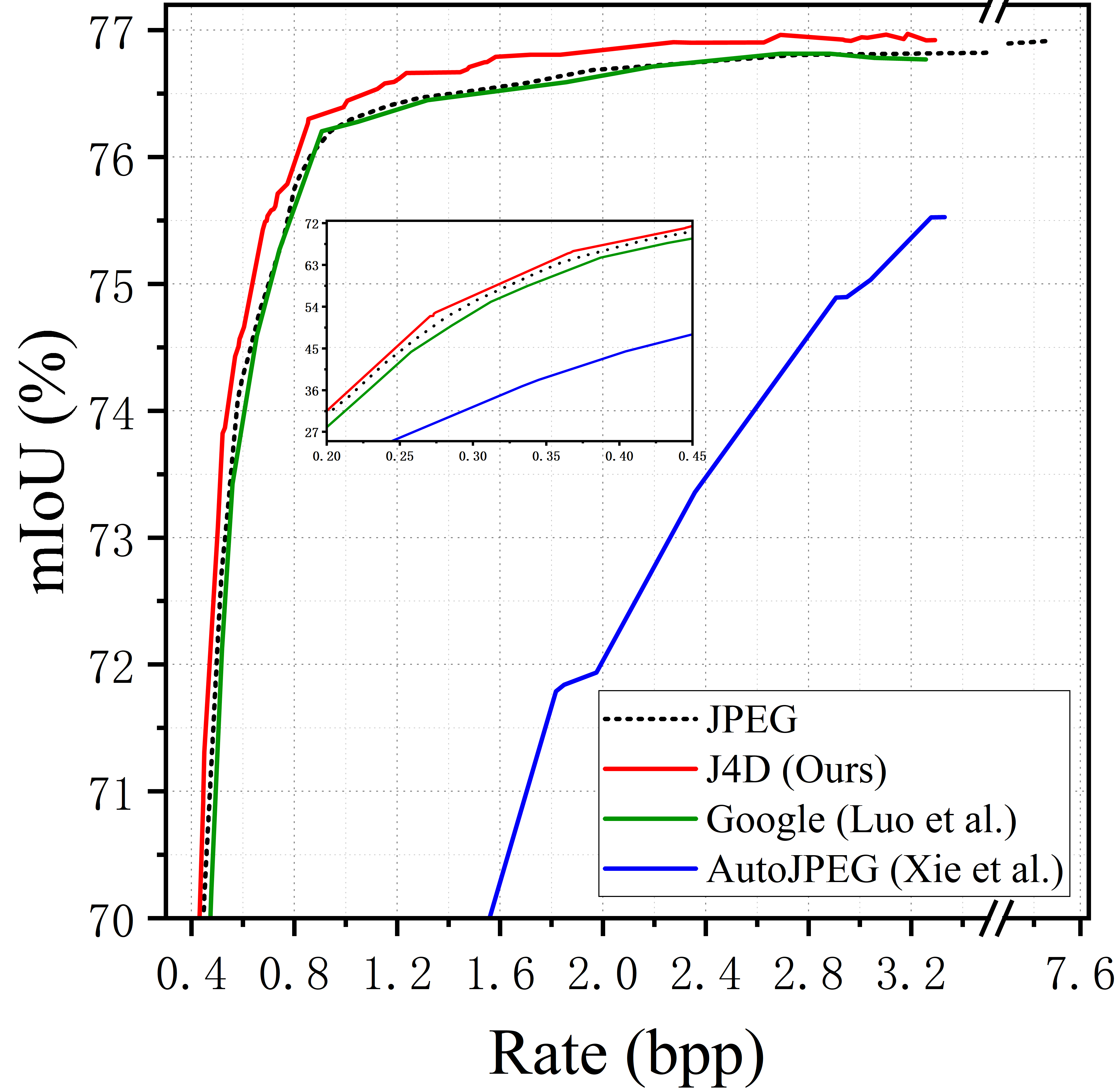}
    \label{fig:VOC_R50}}
    \caption{R-A performance on Pascal VOC 2012 for two pre-trained semantic segmentation models with architectures (a) DeepLabv3-MobileNet and (b) DeepLabv3-ResNet50.}
    \label{fig:VOC}
    \vspace{-3mm}
\end{figure}

\textbf{Results on Pascal VOC 2012}. \cref{fig:VOC} illustrates the superior R-A performance of J4D across two DNNs. Compared to the default JPEG, J4D achieves up to a 79.07\% reduction in rate without accuracy loss or up to a 6.35\% increase in mean intersection-over-union (mIoU) at a fixed rate. 

For further comparison, \cref{tab:table_CUB200,tab:table_ImageNet,tab:table_VOC} in the supplementary material (\cref{sec:tables}) provide numerical summaries of the results presented in the R-A curves, offering a detailed analysis of the performance differences among methods.

\subsection{Analysis}

\begin{figure*}[!ht]
\centering
\subfloat[ResNet18]{\includegraphics[height=0.2\textwidth]{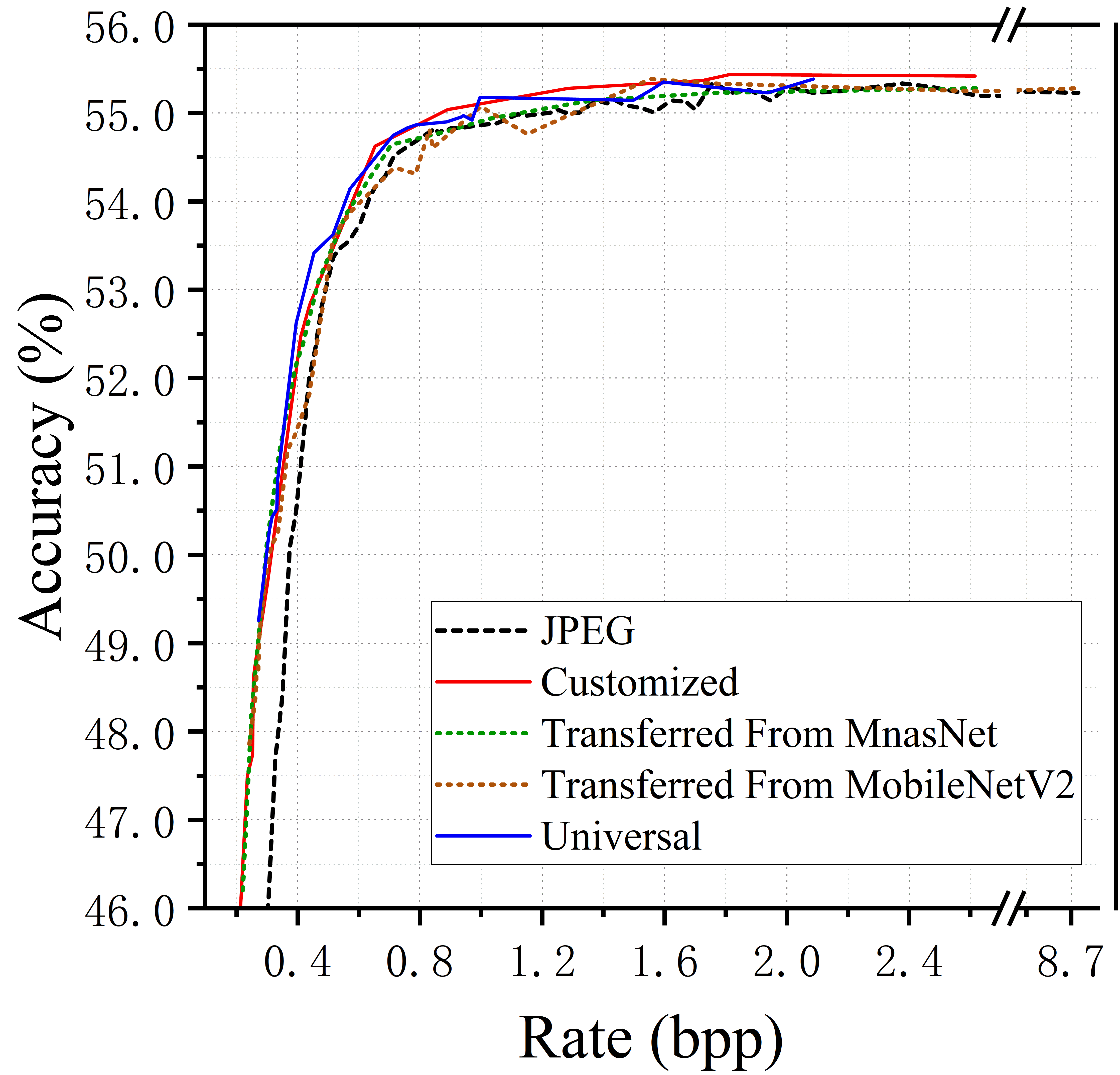}%
\label{fig:Analysis_resnet18}}
\hfil
\subfloat[MobileNetV2]{\includegraphics[height=0.2\textwidth]{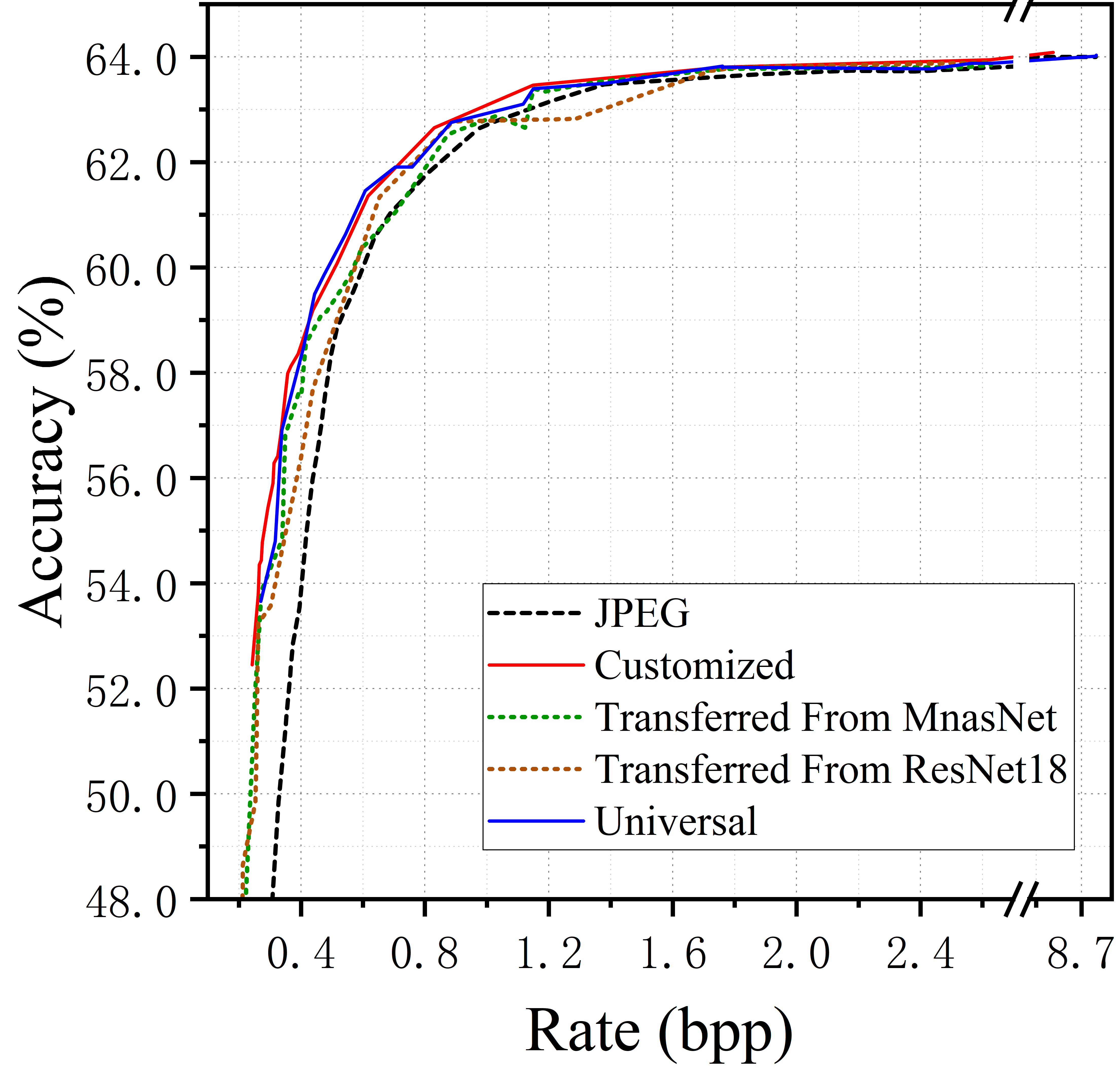}%
\label{fig:Analysis_mobilenet_v2}}
\hfil
\subfloat[MnasNet]{\includegraphics[height=0.2\textwidth]{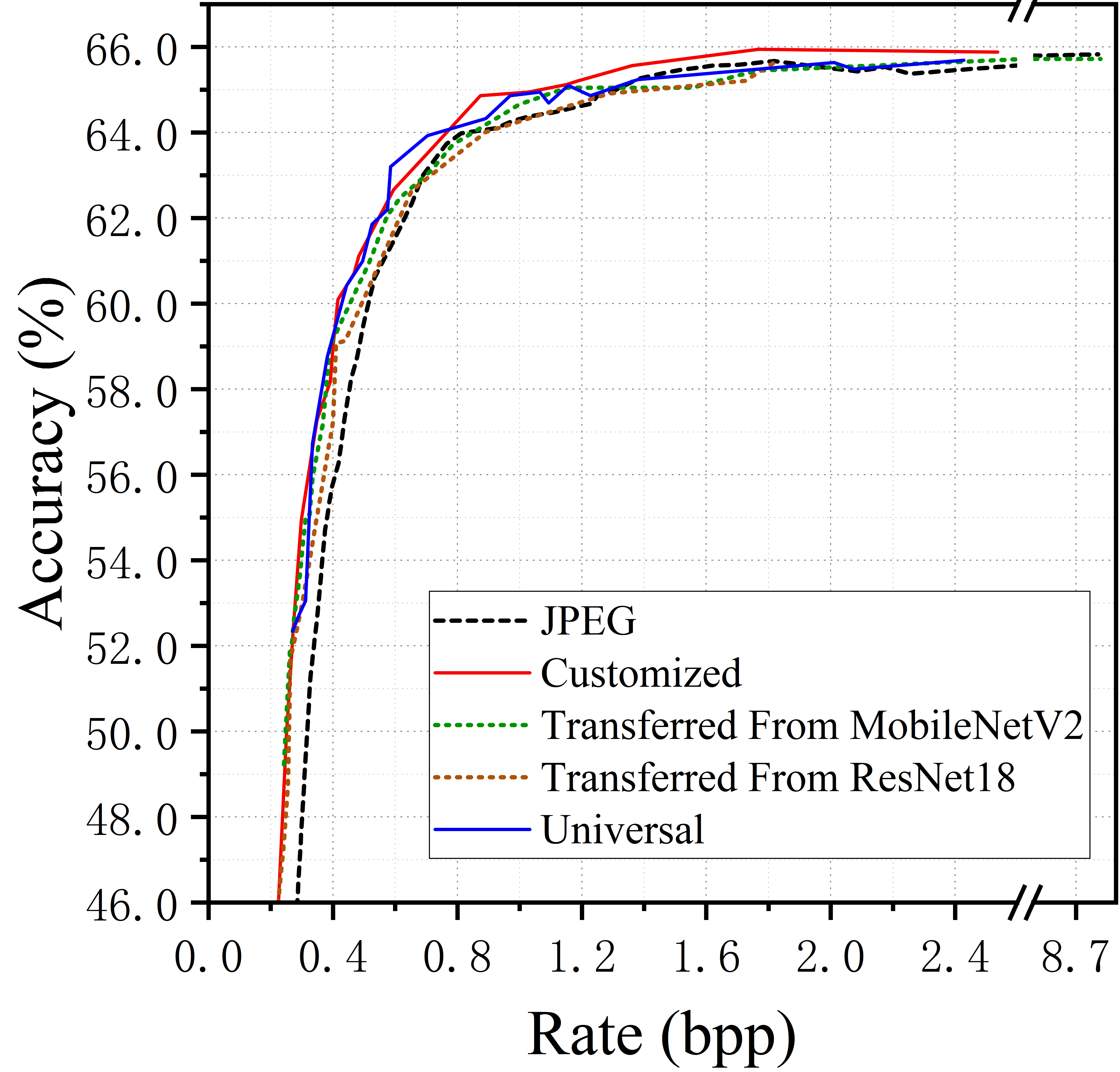}%
\label{fig:Analysis_MnasNet}}
\caption{R-A performance analysis on CUB200 for quantization table transferability and universality.}
\label{fig:Analysis}
\end{figure*}

\textbf{Evaluating Quantization Table Transferability}.
\cref{fig:Analysis} examines the transferability of our optimized quantization tables on CUB200, evaluating the performance of one DNN model when using quantization tables optimized for other DNNs. For instance, \cref{fig:Analysis_resnet18} compares the R-A performance of ResNet18 using its own customized quantization table versus tables optimized for MnasNet and MobileNetV2. Overall, the transferred quantization tables deliver decent results, outperforming JPEG in most cases. 

\begin{figure}[!ht]
    \centering
    \subfloat[ConvNeXt-tiny]{\includegraphics[height=0.4\linewidth]{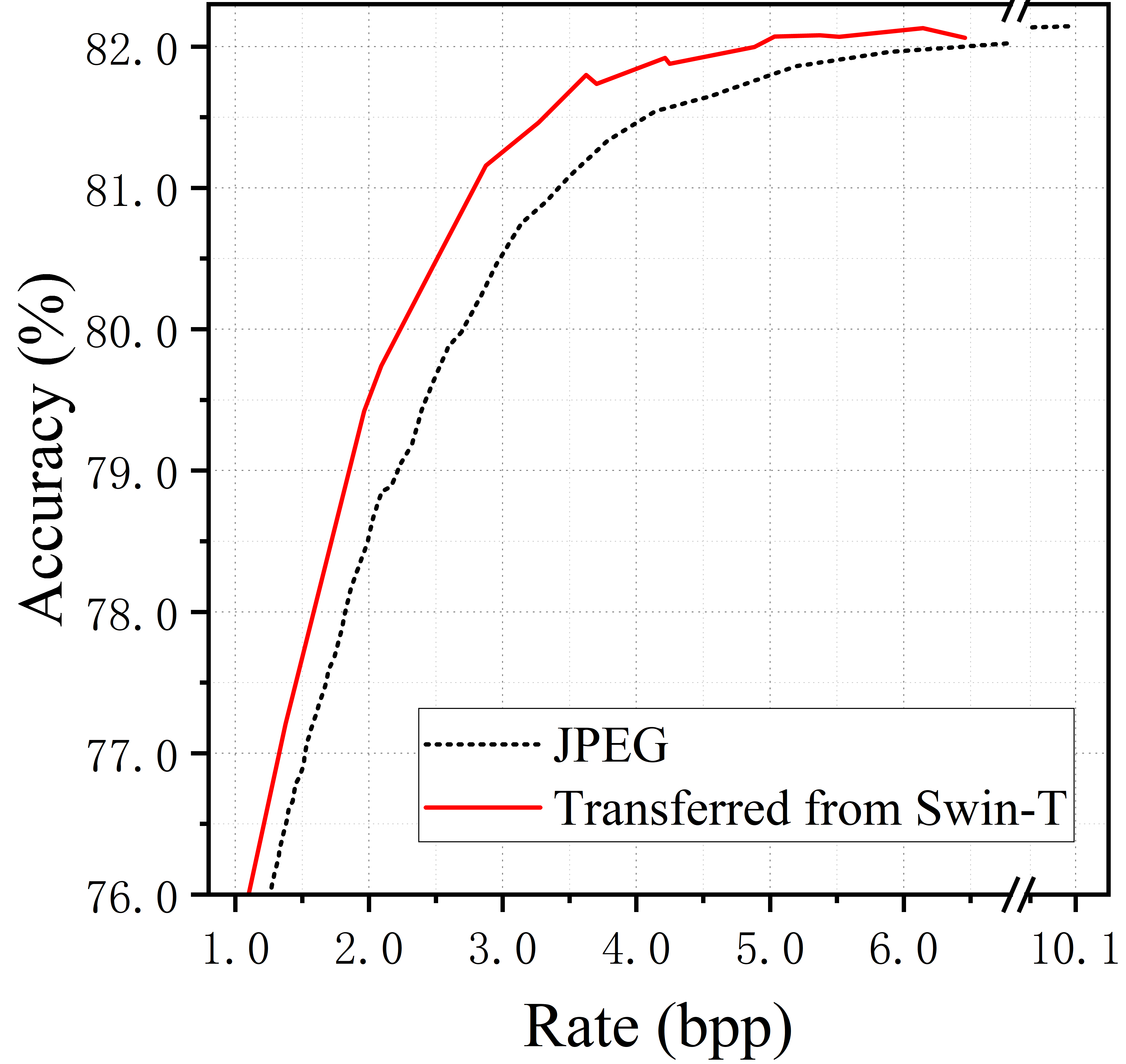}
    \label{fig:trans_cnn}}
    \subfloat[ConvNeXt-tiny ($w_2$)]{\includegraphics[height=0.4\linewidth]{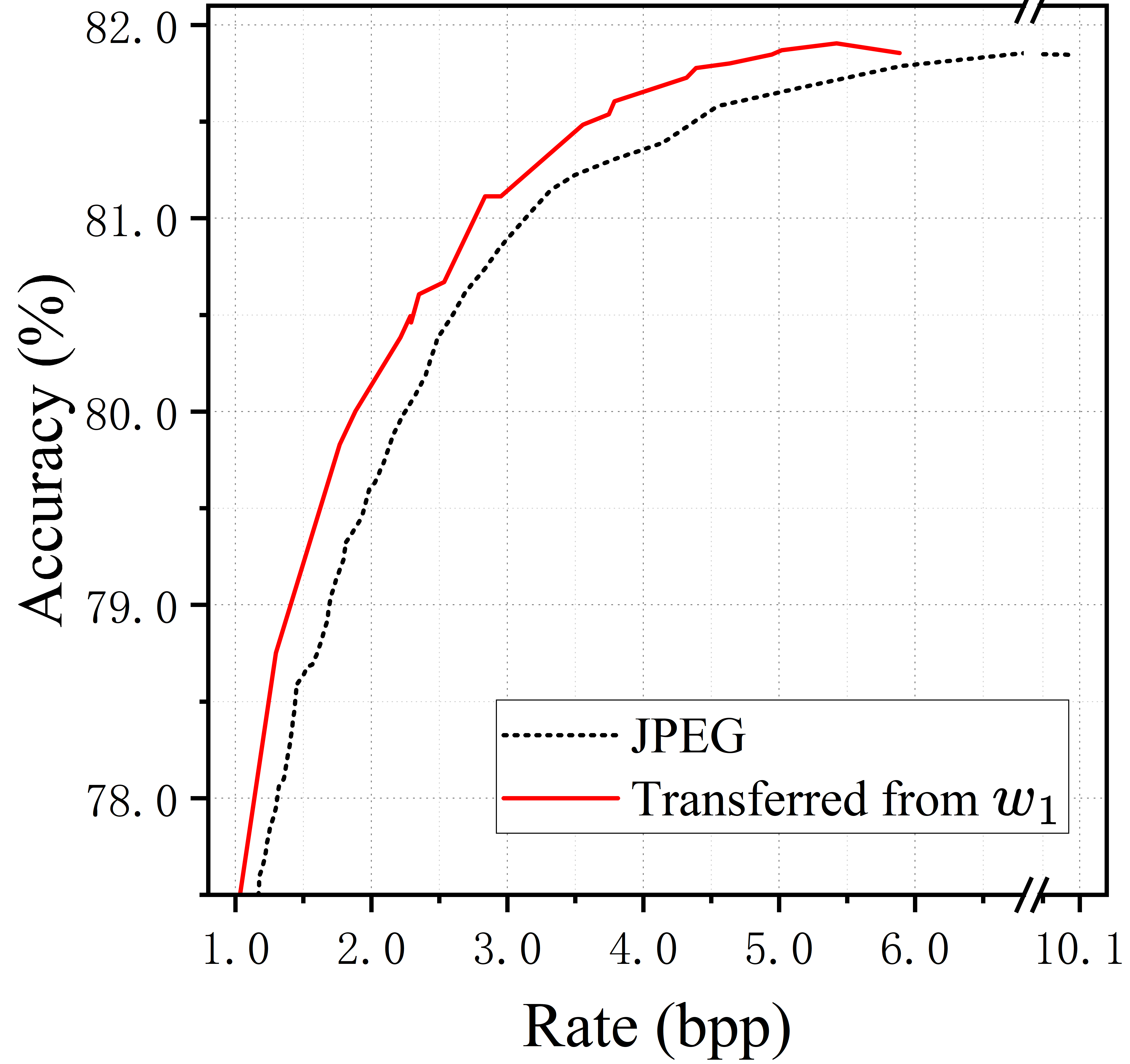}
    \label{fig:weight_transfer}}
    \caption{More analysis on quantization table transferability.
    }
    \label{fig:more_trans}
\end{figure}

Other interesting aspects of transferability are investigated in \cref{fig:more_trans}. In \cref{fig:trans_cnn}, we evaluate the performance of ConvNeXt-tiny on ImageNet-1K when using quantization tables optimized for Swin-T, demonstrating quantization table transferability between convolution-based (ConvNeXt-tiny) and transformer-based (Swin-T) models. In \cref{fig:weight_transfer}, we evaluate the performance of ConvNeXt-tiny with weight $w_2$ when using quantization tables optimized for ConvNeXt-tiny with weight $w_1$, demonstrating quantization table transferability between different pre-trained weights for the same model architecture\footnote{Weight $w_2$ from timm \cite{rw2019timm} is only used in \cref{fig:weight_transfer}. For all other experiments, ConvNeXt-tiny uses weight $w_1$ from PyTorch.}. In both cases, the transferred quantization tables achieve satisfactory results, outperforming JPEG consistently across the entire rate range.

\textbf{Training Universal Quantization Tables}.
While the transferred quantization tables achieve competitive results, the customized tables still demonstrate a nontrivial advantage in R-A performance compared to the transferred ones. To address the limitation of transferring the optimized quantization tables between models, we optimize a set of universal quantization tables for all 3 tested models on CUB200. This is done by replacing the inference loss $\mathcal{L}$ in \cref{eq:final_Total_loss} of a single DNN model with the average inference loss of 3 tested DNN models so that the trained quantization tables take care of the DNN vision for all 3 models. As illustrated in \cref{fig:Analysis}, the universal quantization tables generally achieve comparable R-A performance to that of customized quantization tables for each individual model, showing their strong generalization to different DNNs.

\begin{figure}[!ht]
    \centering
    \subfloat[Luminance]{\includegraphics[height=0.4\linewidth]{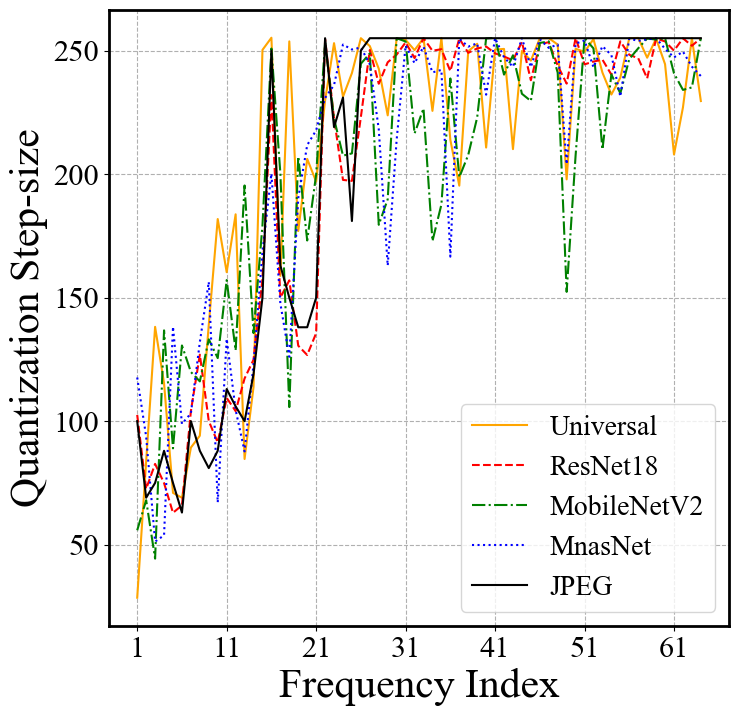}
    \label{fig:Q_table_Luma}}
    \subfloat[Chrominance]{\includegraphics[height=0.4\linewidth]{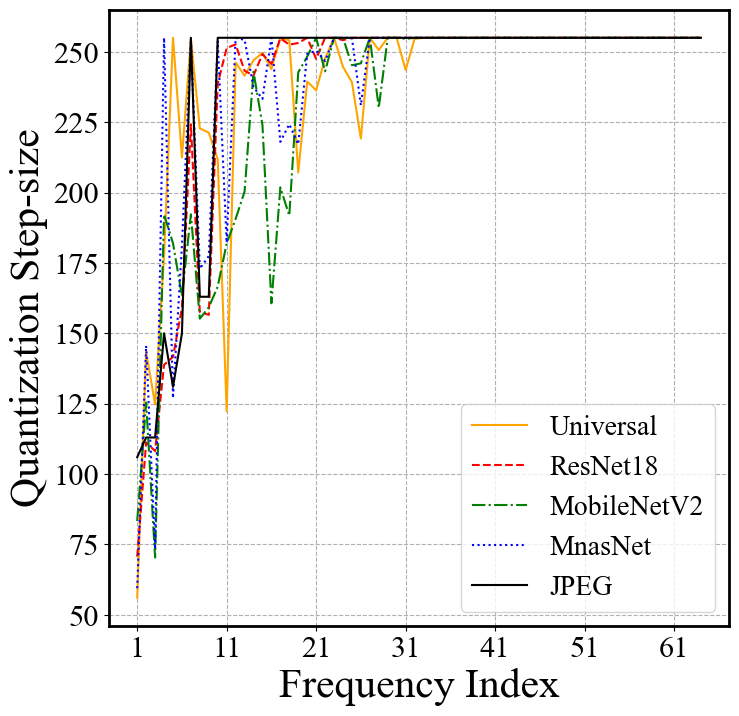}
    \label{fig:Q_table_Chroma}}
    \caption{The optimized luminance and chrominance quantization tables on CUB200, flattened in the zigzag order, are shown in \cref{fig:Q_table_Luma,fig:Q_table_Chroma}, respectively, at the same rate of 0.3 bpp.}
    \label{fig:Q_table}
\end{figure}

\begin{figure}[!ht]
    \centering
    \includegraphics[width=0.5\linewidth]{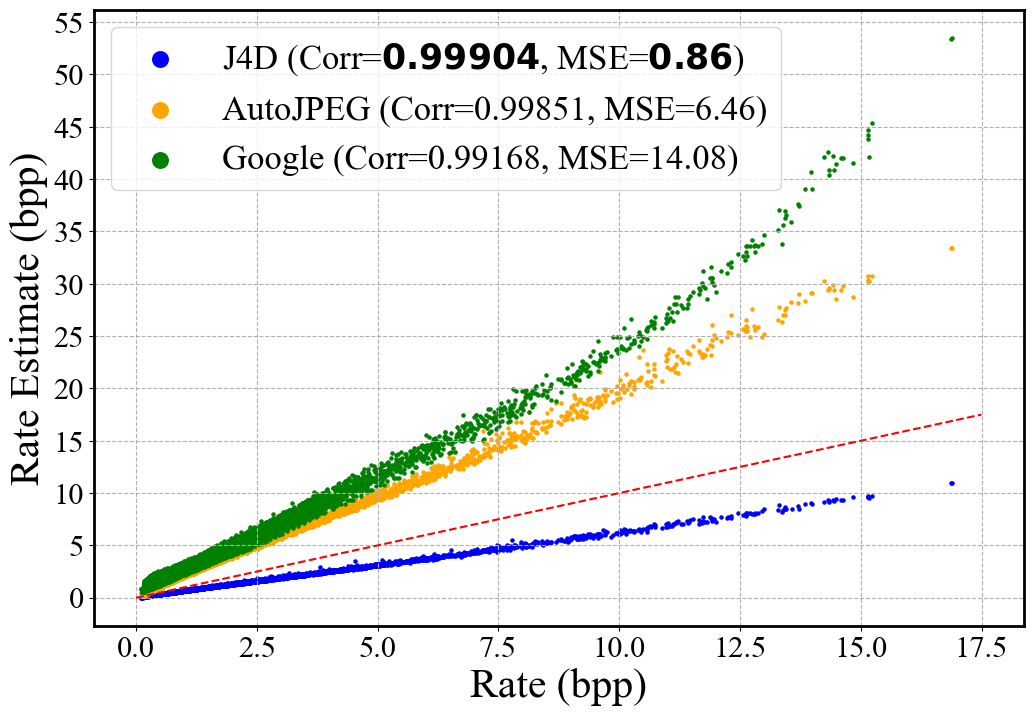}
    \caption{Comparison of $(\mathcal{R}, \mathcal{\hat{R}})$ pairs for different rate estimators, including those used in J4D, Google, and AutoJPEG.}
    \label{fig:correlation}
\end{figure}

\textbf{Comparison of Rate Estimators}.
We compare the performance of our rate estimator to those of Google and AutoJPEG using the Pearson correlation coefficient and mean squared error (MSE) between actual compression rates and rate estimates. To conduct this evaluation, in \cref{fig:correlation}, we randomly sampled 100 images from the ImageNet-1K validation set and generated the actual rates using the default JPEG quantization tables with various quality factors ranging from 1 to 100. From \cref{fig:correlation}, it's clear that the $(\mathcal{R}, \mathcal{\hat{R}})$ pairs resulting from J4D shows a stronger linear behavior than the other two methods, and they are significantly closer to the diagonal (shown as a red dashed line) than those of the other two methods. These visual impressions are further quantified by the correlation and MSE values shown in the legend of \cref{fig:correlation}. Both the largest correlation and the smallest MSE are achieved by J4D, validating that our rate estimator provides a more accurate approximation of the actual rate compared to the benchmark methods.

\textbf{Comparison of Quantization Tables}.
In \cref{fig:Q_table_Luma,fig:Q_table_Chroma}, we compare the universal and customized quantization tables for 3 models used on CUB200 together with those in the default JPEG at a fixed rate.
The analysis reveals that DNN models show different perceptions compared to humans, favoring the fidelity of higher frequencies while allowing lower frequencies to be compressed more aggressively, reflected by generally smaller quantization steps at high frequencies and larger quantization steps at low frequencies of the DNN-oriented quantization tables compared to the default JPEG ones. Besides, the figures also show that the universal quantization tables, designed to accommodate the characteristics of all three models, generally take quantization step sizes less extreme than those of customized quantization tables.

Please refer to the supplementary material (\cref{sec:jp_ju,sec:data_trans,sec:r_d,sec:visual_images}) for more analysis on visual quality of images compressed by JPEG and J4D, quantization table transferability between datasets, and the rate-distortion (R-D) performance of J4D, etc.


\section{Conclusion} \label{sec:Conclusion}


This paper introduces J4D, a novel approach to optimize JPEG compression specifically for DNN inference. By introducing a differentiable JPEG codec and an entropy-based rate estimator, J4D effectively solves the rate-accuracy optimization problem by backpropagation-based gradient descent. Extensive experiments demonstrate that J4D outperforms the default JPEG for human vision and other DNN-oriented JPEG codecs in the R-A sense, making it the SOTA JPEG compression for DNN vision. Moreover, the study, for the first time, highlights the potential for designing universal JPEG quantization tables suitable for all kinds of DNN architectures.

{
    \small
    \bibliographystyle{ieeenat_fullname}
    \bibliography{main}
}

\clearpage
\setcounter{page}{1}
\maketitlesupplementary


\section{Visual Comparison of Compressed Images} \label{sec:visual_images}

\begin{figure}[!ht]
\centering
\begin{subfigure}[b]{0.42\textwidth}
\centering 
\includegraphics[width=\linewidth]{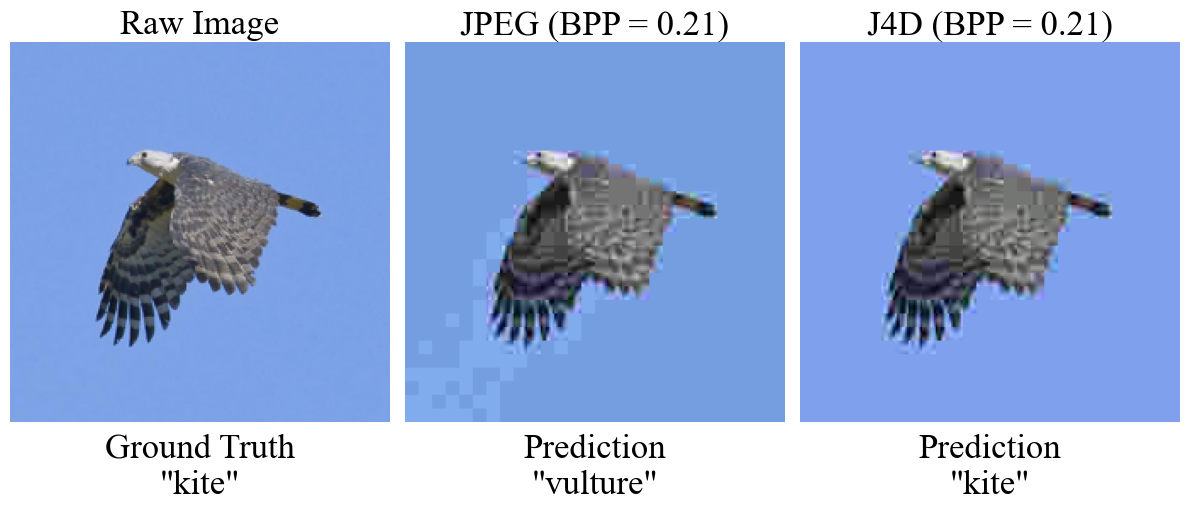}
        \vspace{-2ex}
        \label{refig:low_rate_1}
\includegraphics[width=\linewidth]{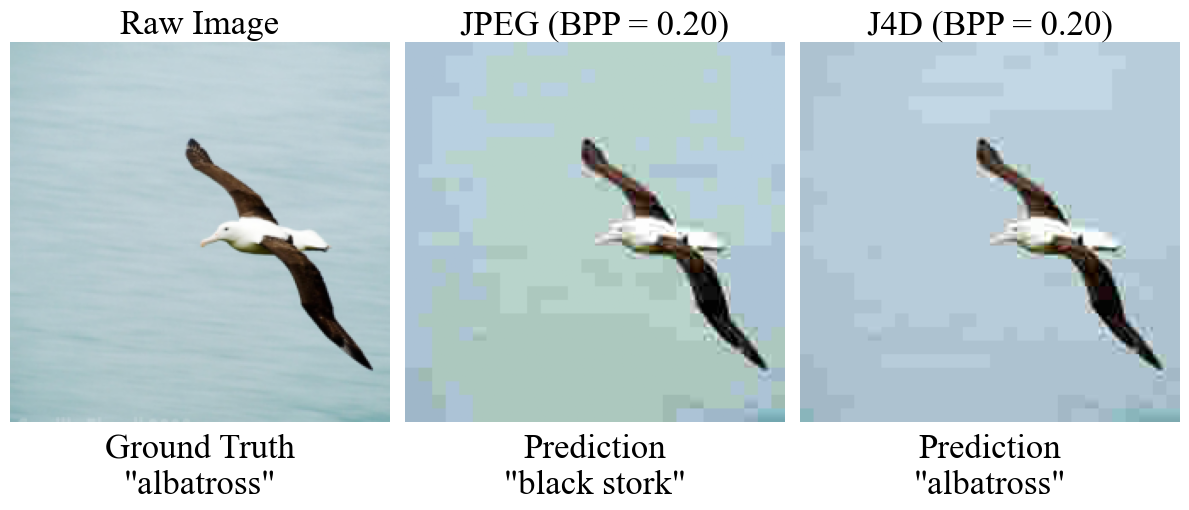}
        \caption{MobileNetV2 (ImageNet-1K)}
        \vspace{+1ex}
        \label{refig:low_rate_2}

\includegraphics[width=\linewidth]{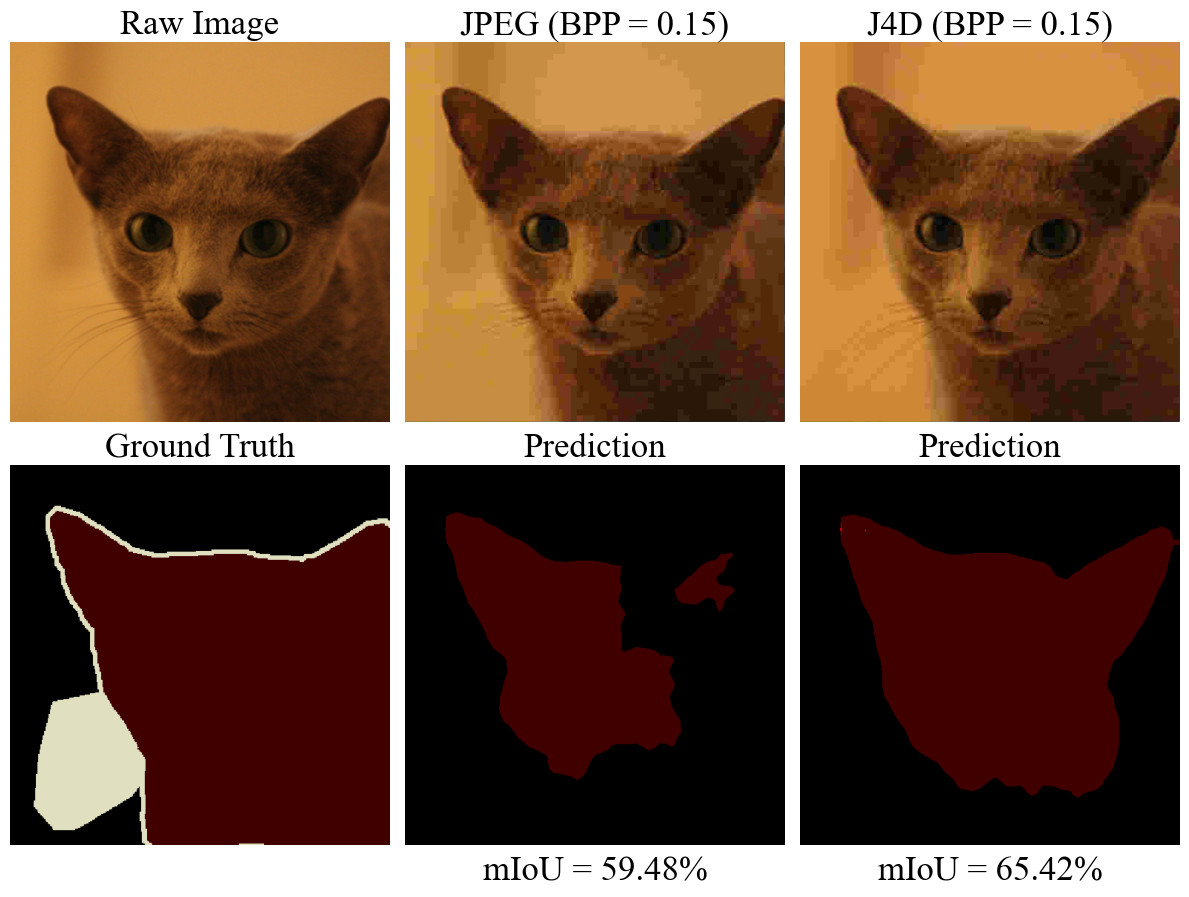}
        \vspace{-2ex}
        \label{refig:seg_low_rate_1}
\includegraphics[width=\linewidth]{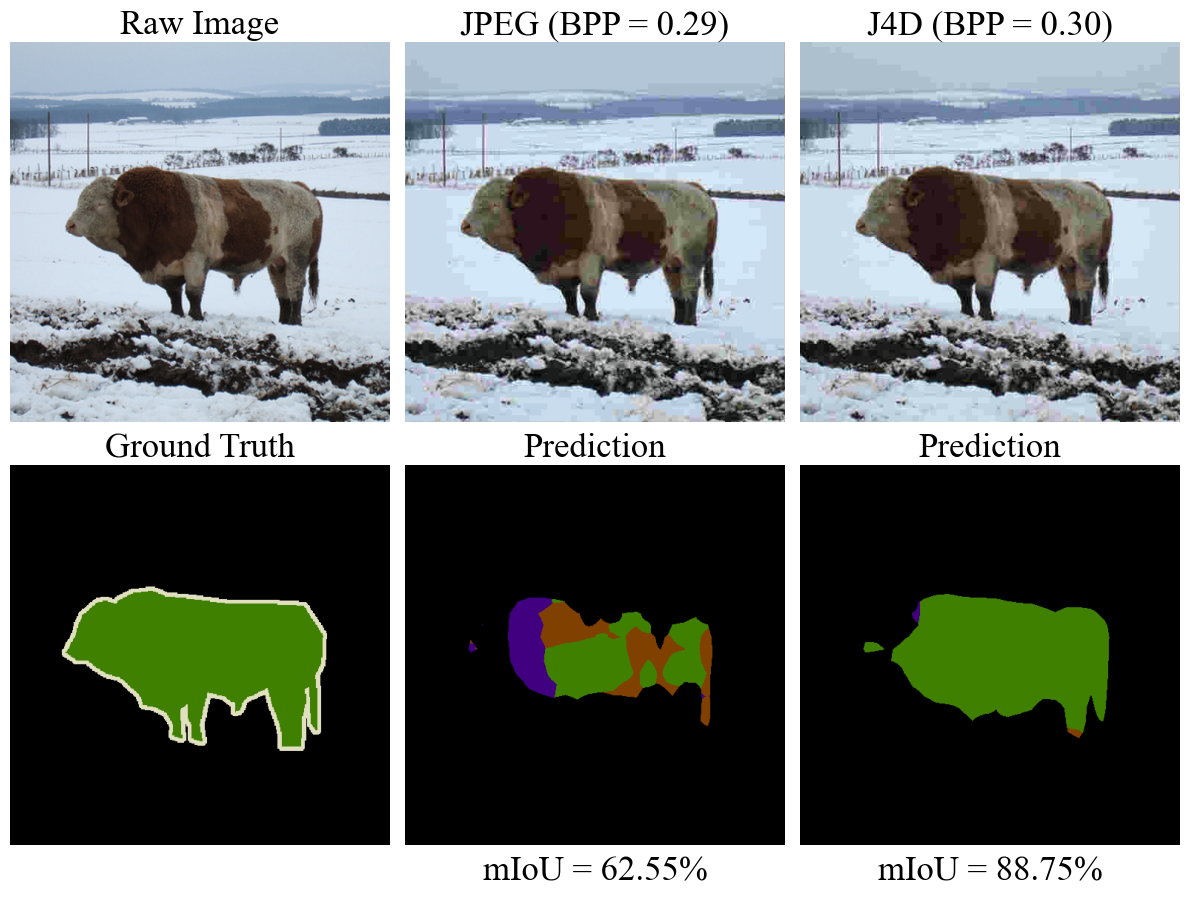}
        \caption{DeepLabv3-MobileNet (Pascal VOC 2012)}
        \label{refig:seg_low_rate_2}
\end{subfigure}
\caption{Visual comparison of compressed images resulting from the default JPEG and J4D.} 
\label{fig:visual}
\end{figure}

In \cref{fig:visual}, we illustrate some images compressed by the default JPEG and J4D at the same bit rate, together with the raw images provided as references. In \cref{refig:low_rate_2}, we present two examples from the ImageNet-1K validation set. It is clear from \cref{refig:low_rate_2} that the images compressed by J4D preserve more edge (high frequency) information and/or have better contrast with the background. They are correctly classified by MobileNetV2, whereas images compressed by the default JPEG are misclassified. In \cref{refig:seg_low_rate_2}, we show two other examples from the Pascal VOC 2012 validation set, where J4D achieves much better segmentation results than the default JPEG, with DeepLabv3-MobileNet serving as the target DNN. Again, the same observation about the high frequency information preservation can be made as well. 

\section{Numerical Result Comparison} \label{sec:tables}

Corresponding to \cref{fig:CUB200}, \cref{fig:ImageNet}, and \cref{fig:VOC},  \cref{tab:table_CUB200,tab:table_ImageNet,tab:table_VOC} show some specific numerical results for CUB200, ImageNet-1K, and Pascal VOC 2012, respectively. For all tested pre-trained models, these tables show a more detailed analysis of the performance differences between each DNN-oriented JPEG codec and the default JPEG at a fixed rate or a fixed accuracy. Again, the performance gains provided by J4D are significant and consistent across datasets, models, and tasks. 

\begin{table*}[!ht]
\centering
\renewcommand{\arraystretch}{0.6} 
\resizebox{\linewidth}{!}{
\begin{tabular}{l|cc|cc|cc|cc}
\toprule
\multirow{2}{*}{Pre-Trained Model} &\multicolumn{2}{c|}{JPEG} &\multicolumn{2}{c}{J4D} &\multicolumn{2}{c}{Google} &\multicolumn{2}{c}{AutoJPEG} \\
& Rate (bpp) & Acc (\%) & Rate (bpp) & Acc (\%)  & Rate (bpp) & Acc (\%) & Rate (bpp) & Acc (\%)\\ 
\midrule
\multirow{2}{*}{ResNet18}    
& 0.22 & 34.92 & 0.21 & 45.96 (+11.05) & 0.22 & 36.04 (+1.12) & 0.24 & 27.17 (-7.75) \\
& 8.74 & 55.23 & 1.74 (-80.05\%) & 55.25 & 4.75 (-45.63\%) & 55.26 & 7.29 (-16.63\%) & 55.23 \\
\cmidrule{1-9}
\multirow{2}{*}{MobileNetV2} 
& 0.25 & 40.85 & 0.24 & 52.45 (+11.60) & 0.25 & 39.92 (-0.93) & 0.24 & 27.48 (-13.38) \\
& 8.74 & 63.40 & 2.62 (-69.97\%) & 63.95 & 4.78 (-45.34\%) & 63.92 & 6.80 (-22.20\%) & 64.01 \\
\cmidrule{1-9}
\multirow{2}{*}{MnasNet}     
& 0.25 & 40.97 & 0.25 & 49.95 (+8.97) & 0.24 & 39.85 (-1.12) & 0.27 & 31.05 (-9.92)\\
& 8.74 & 65.83 & 1.75 (-79.94\%) & 65.83 & 4.73 (-45.86\%) & 65.91 & 7.38 (-15.60\%) & 65.81 \\
\bottomrule
\end{tabular}
}
\caption{R-A performance summary, derived from \cref{fig:CUB200}, enables a numerical comparison between all tested methods, covering both the same rate and the same accuracy comparison. The values shown in parentheses represent the gain in accuracy at a fixed rate or the percentage of reduction in rate at fixed accuracy in comparison to the default JPEG.}
\label{tab:table_CUB200}
\end{table*}


\begin{table*}[!ht]
\centering
\renewcommand{\arraystretch}{0.6} 
\resizebox{\linewidth}{!}{
\begin{tabular}{l|cc|cc|cc|cc}
\toprule
\multirow{2}{*}{Pre-Trained Model} &\multicolumn{2}{c|}{JPEG} &\multicolumn{2}{c}{J4D} &\multicolumn{2}{c}{Google} &\multicolumn{2}{c}{AutoJPEG} \\
& Rate (bpp) & Acc (\%) & Rate (bpp) & Acc (\%)  & Rate (bpp) & Acc (\%) & Rate (bpp) & Acc (\%)\\ 
\midrule
\multirow{2}{*}{MobileNetV2}   
& 2.02 & 69.54 & 2.00 & 70.42 (+0.88) & 1.95 & 69.81 (+0.27) & 1.93 & 67.47 (-2.07) \\
& 10.10 & 71.86 & 5.15 (-49.02\%) & 71.86 & 6.65 (-34.18\%) & 71.88 & 6.78 (-32.91\%) & 71.80 \\
\cmidrule{1-9}
\multirow{2}{*}{MnasNet}       
& 1.93 & 71.20 & 1.92 & 71.98 (+0.79) & 1.96 & 71.65 (+0.46) & 2.00 & 69.10 (-2.09) \\
& 10.10 & 73.41 & 4.07 (-59.71\%) & 73.42 & 6.71 (-33.60\%) & 73.37 & 7.78 (-23.03\%) & 73.39 \\
\cmidrule{1-9}
\multirow{2}{*}{ConvNeXt-tiny} 
& 1.86 & 78.16 & 1.88 & 79.54 (+1.38) & 1.88 & 79.04 (+0.88) & 1.93 & 78.93 (+0.77) \\
& 10.10 & 82.14 & 5.88 (-41.77\%) & 82.17 & 6.56 (-35.06\%) & 82.04 & 7.21 (-28.64\%) & 82.12 \\
\cmidrule{1-9}
\multirow{2}{*}{Swin-T}        
& 1.99 & 75.61 & 1.96 & 77.82 (+2.21) & 1.92 & 76.78 (+1.17) & 2.00 & 76.17 (+0.55) \\
& 10.10 & 81.13 & 5.04 (-50.13\%) & 81.14 & 6.60 (-34.64\%) & 81.07 & 8.47 (-16.16\%) & 81.10 \\
\bottomrule
\end{tabular}
}
\caption{R-A performance summary, derived from \cref{fig:ImageNet}.}
\label{tab:table_ImageNet}
\end{table*}


\begin{table*}[!ht]
\centering
\renewcommand{\arraystretch}{0.6} 
\resizebox{\linewidth}{!}{
\begin{tabular}{l|cc|cc|cc|cc}
\toprule
\multirow{2}{*}{Pre-Trained Model} &\multicolumn{2}{c|}{JPEG} &\multicolumn{2}{c}{J4D} &\multicolumn{2}{c}{Google} &\multicolumn{2}{c}{AutoJPEG} \\
& Rate (bpp) & mIoU (\%) & Rate (bpp) & mIoU (\%)  & Rate (bpp) & mIoU (\%) & Rate (bpp) & mIoU (\%)\\ 
\midrule
\multirow{2}{*}{DeepLabv3-MobileNet} 
& 0.26 & 39.86 & 0.27 & 46.21 (+6.35) & 0.27 & 36.41 (-3.45) & 0.29 & 23.43 (-16.43) \\
& 7.55 & 70.07 & 1.58 (-79.07\%) & 70.08 & 2.83 (-62.52\%) & 70.11 & 6.19 (-18.01\%) & 70.00 \\
\cmidrule{1-9}
\multirow{2}{*}{DeepLabv3-ResNet50} 
& 0.26 & 46.99 & 0.27 & 52.73 (+5.74) & 0.26 & 44.26 (-2.73) & 0.33 & 36.80 (-10.19) \\
& 7.55 & 76.91 & 2.27 (-69.93\%) & 76.91 & 3.82 (-49.40\%) & 76.80 & 5.72 (-24.24\%) & 76.75 \\
\bottomrule
\end{tabular}
}
\caption{R-A performance summary, derived from \cref{fig:VOC}.}
\label{tab:table_VOC}
\end{table*}

\section{Discussion on the Initialization of \texorpdfstring{$\alpha$}{alpha} Tables} \label{sec:alpha_selection}

\begin{figure}[!ht]
\centering
\includegraphics[width=0.4\textwidth]{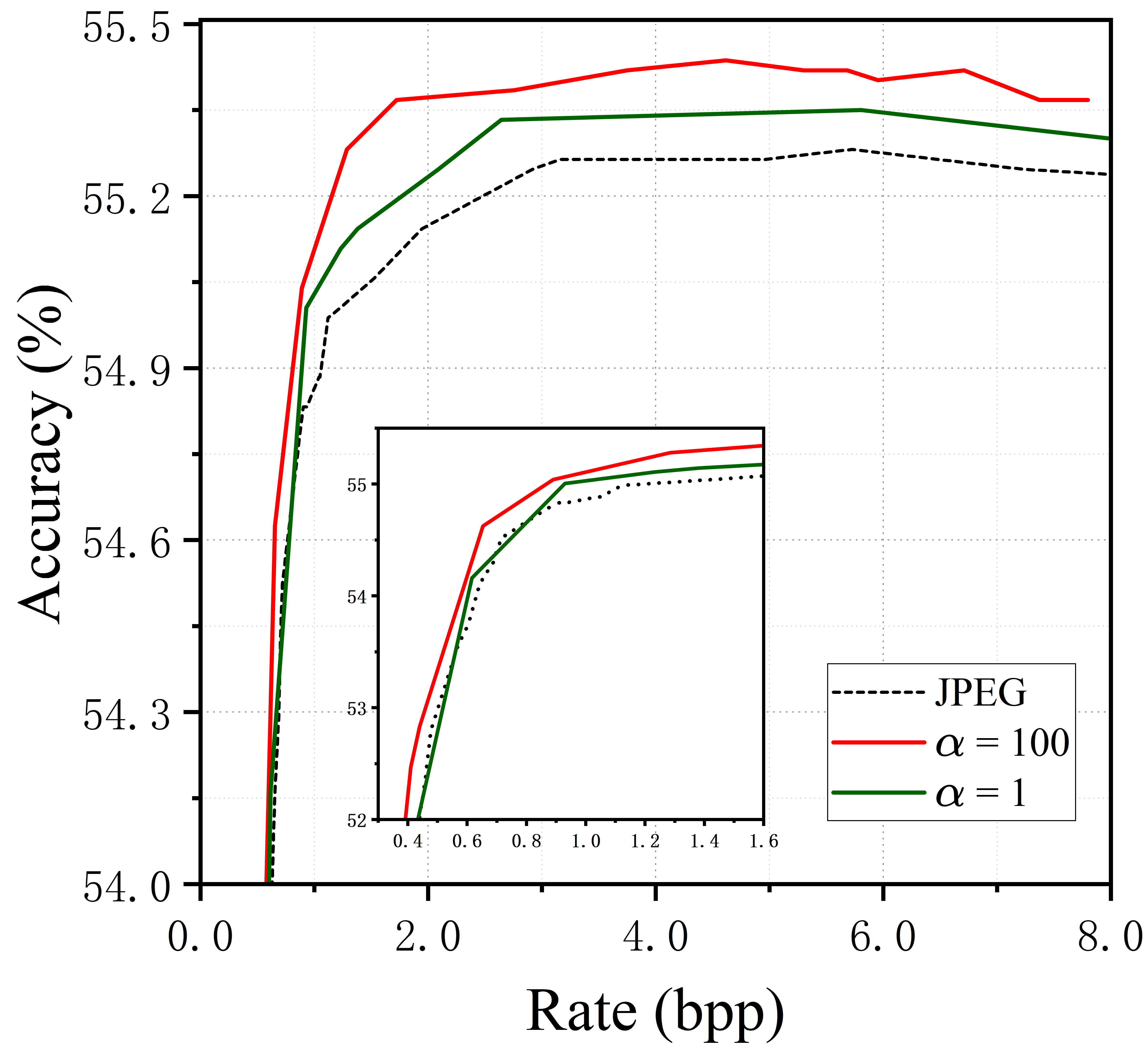}
\caption{The R-A performance of ResNet18 on compressed CUB200 by J4D with  $\alpha = 1$ vs $\alpha = 100$.}
\label{fig:alpha_1_100}
\end{figure}

When using a smaller $\alpha$ value, the CPMF of a probabilistic quantizer is smoother, so there will be a higher probability for a DCT coefficient to be quantized by $\mathcal{Q}_p$ to a reconstruction level further to the right or left, leading to a larger quantization error. Moreover, since the CPMF, thus the MPMF, becomes smoother, the entropy of quantized DCT coefficients gets larger, and so is the actual compression rate. As a result, using a smaller $\alpha$ in $\mathcal{Q}_p$ gives strictly worse R-A performance. This is also verified by our experiments. As shown in \cref{fig:alpha_1_100}, the R-A performance resulting from our training framework with $\alpha = 1$ is consistently worse than that with $\alpha = 100$. Therefore, we prefer to initialize $\alpha$ tables with sufficiently large values to guarantee good R-A performance.

\section{Asymptotic Convergence of Differentiable Soft Quantizers to the Uniform Quantizer} \label{sec:sup_quantier}

\begin{figure}[!ht]
\centering
\includegraphics[width=0.35\textwidth]{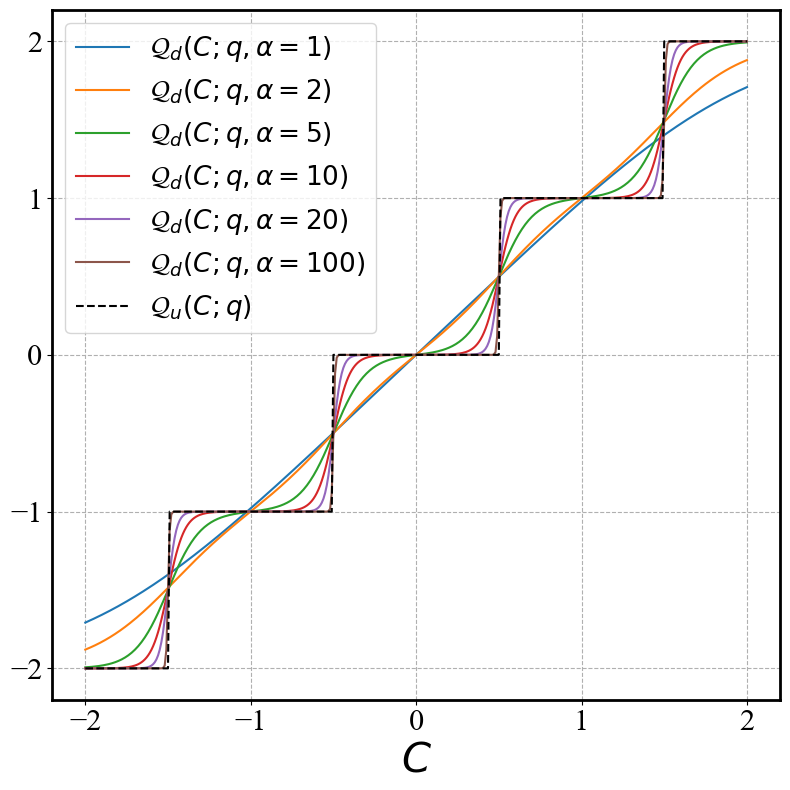}
\caption{An illustration of $\mathcal{Q}_u$ \vs $\mathcal{Q}_d$ with $\alpha$ = $1, 2, 5, 10, 20, 100$, where $L$ and $q$ are set to 2 and 1, respectively.}
\label{fig:Q_u_Q_d}
\end{figure}

As illustrated in \cref{fig:Q_u_Q_d}, $\mathcal{Q}_d$ converges to $\mathcal{Q}_u$ as $\alpha$ increases.

\section{Sensitivity-based Quantization Table Initialization} \label{sec:sup_sens}

Given a pre-trained model, denote $S^{l}_{m}$ as the sensitivity of the model to DCT frequencies as described in \cite{Zheng2023jpegcompliant, Salamah2024jpegcompliant}, where $l=1,2,3$ corresponds to the color channels Y, Cb and Cr, respectively, and $1 \leq m \leq M$ represents the frequency position. We initialize each quantization step based on the reciprocal of the sensitivity value for each DCT frequency as follows:
\begin{align}
    \tau &= \max_{m}\left\{\frac{1}{S^{1}_{m}}\right\}, \\
    q_m &= \frac{\beta}{S^{1}_{m} \cdot \tau}, \\
    q_{M+m} &= \frac{2 \beta}{(S^{2}_{m} + S^{3}_{m})\cdot \tau}, 
\end{align}
where $1 \leq m \leq M$, $\tau$ is an intermediate variable for normalization, and $\beta$ is a scaling factor as mentioned in \cref{sec:Exp_setting}.

\begin{figure}[!ht]
    \centering
    \subfloat[ResNet18]{\includegraphics[width=\linewidth,keepaspectratio]{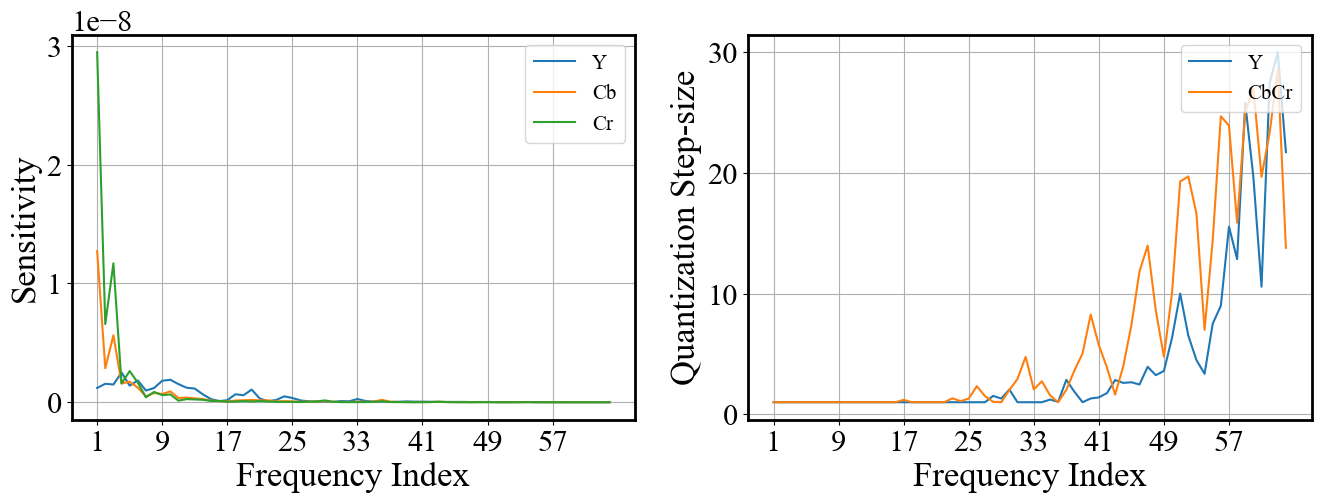}}
    \hfill
    \subfloat[MobileNetV2]{\includegraphics[width=\linewidth,keepaspectratio]{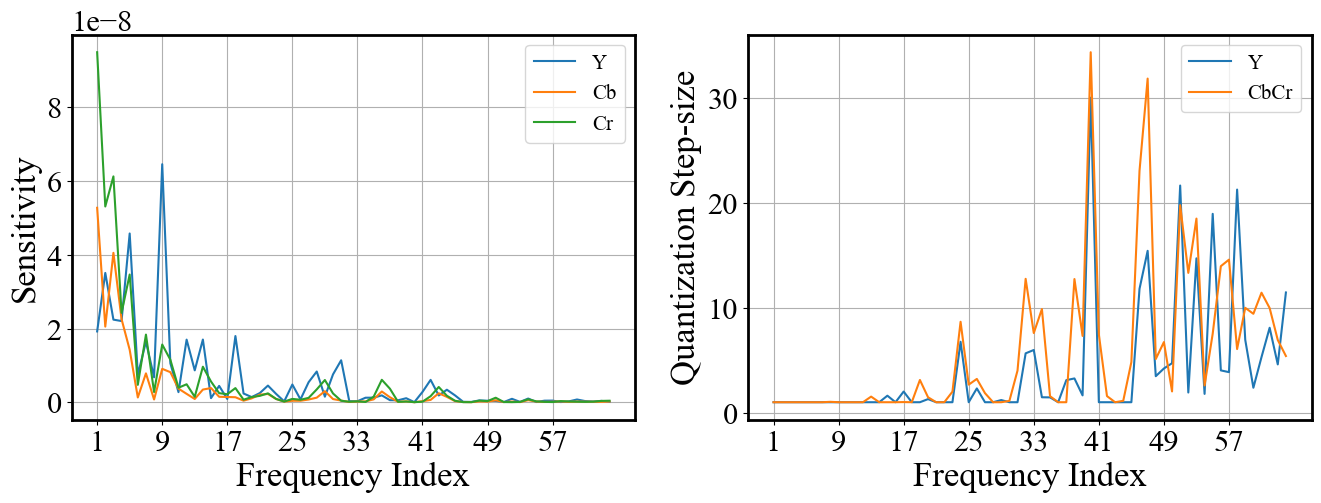}}
    \hfill
    \subfloat[MnasNet]{\includegraphics[width=\linewidth,keepaspectratio]{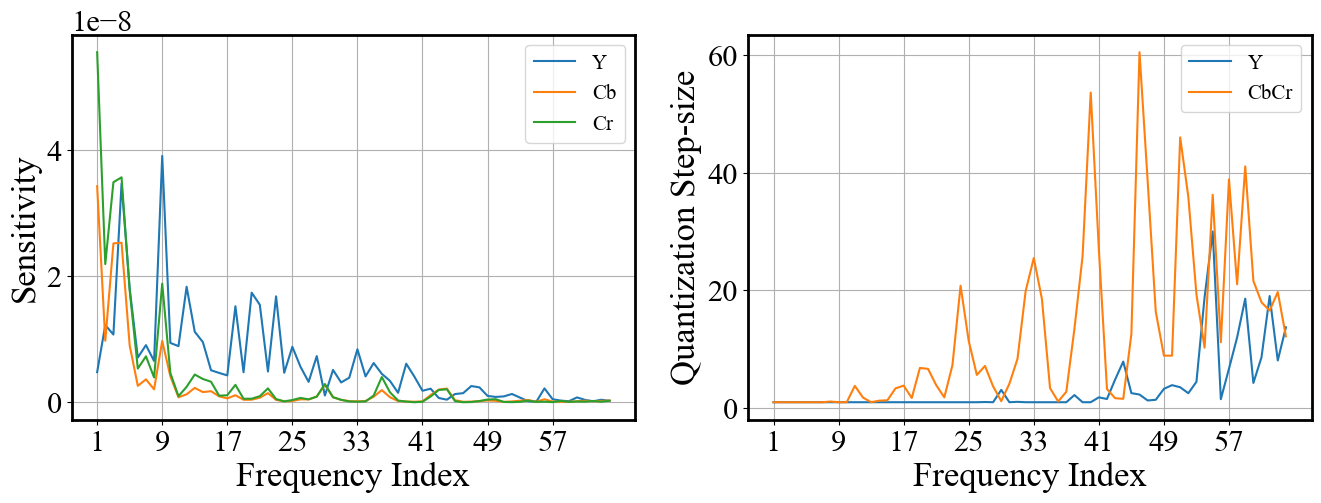}}
    \caption{The sensitivity values (left) and the corresponding sensitivity-based initial quantization tables (right) on CUB200.}
    \label{fig:sensitivity_cub200}
\end{figure}

\begin{figure}[!ht]
    \centering
    \subfloat[MobileNetV2]{\includegraphics[width=\linewidth,keepaspectratio]{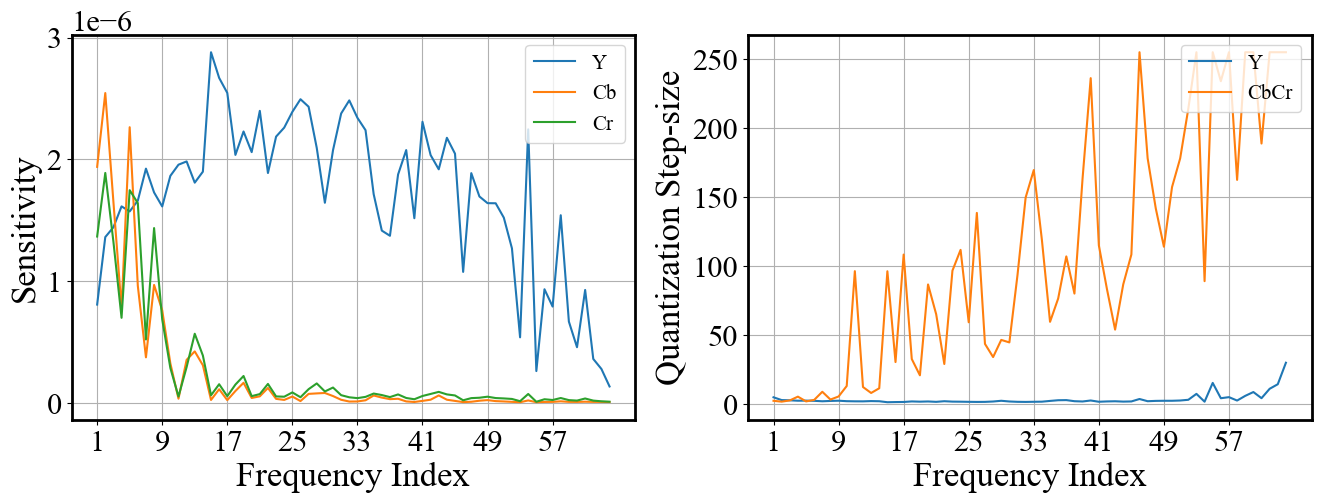}}
    \hfill
    \subfloat[MnasNet]{\includegraphics[width=\linewidth,keepaspectratio]{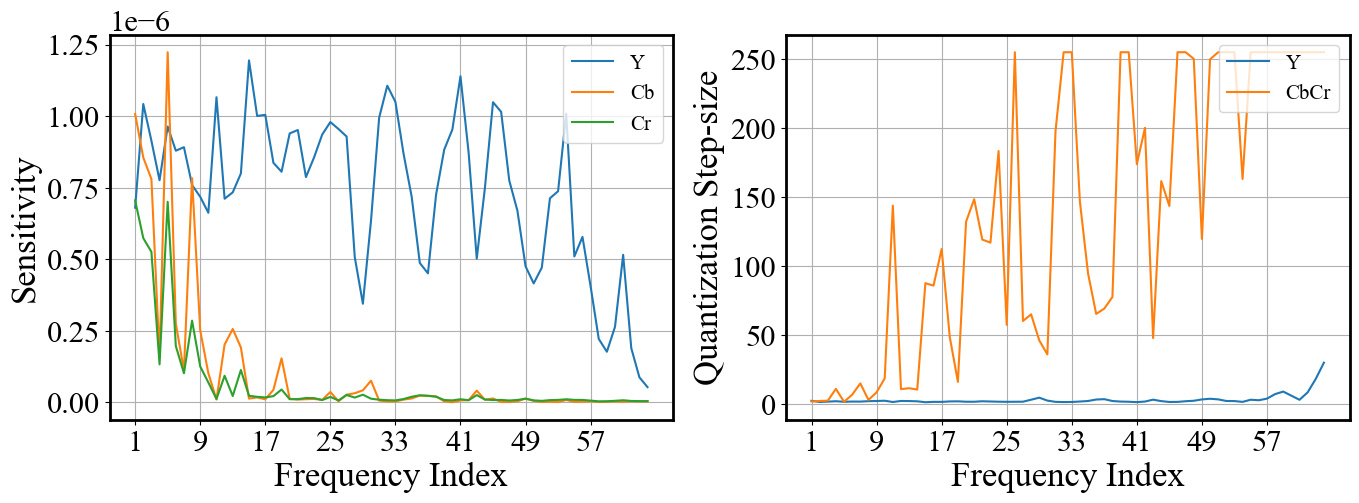}}
    \hfill
    \subfloat[ConvNeXt-tiny]{\includegraphics[width=\linewidth,keepaspectratio]{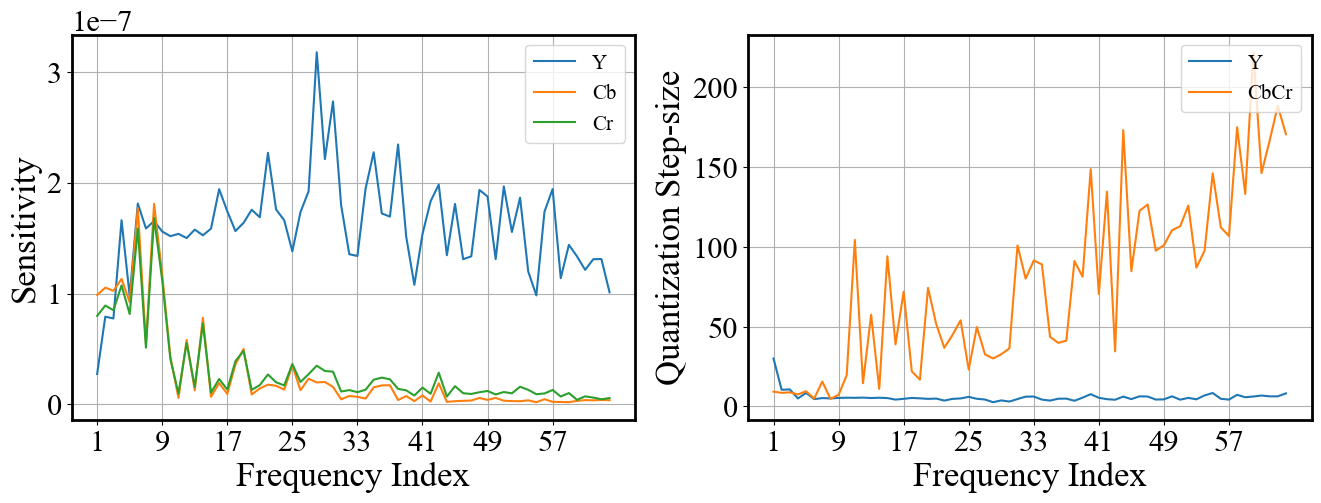}}
    \hfill
    \subfloat[Swin-T]{\includegraphics[width=\linewidth,keepaspectratio]{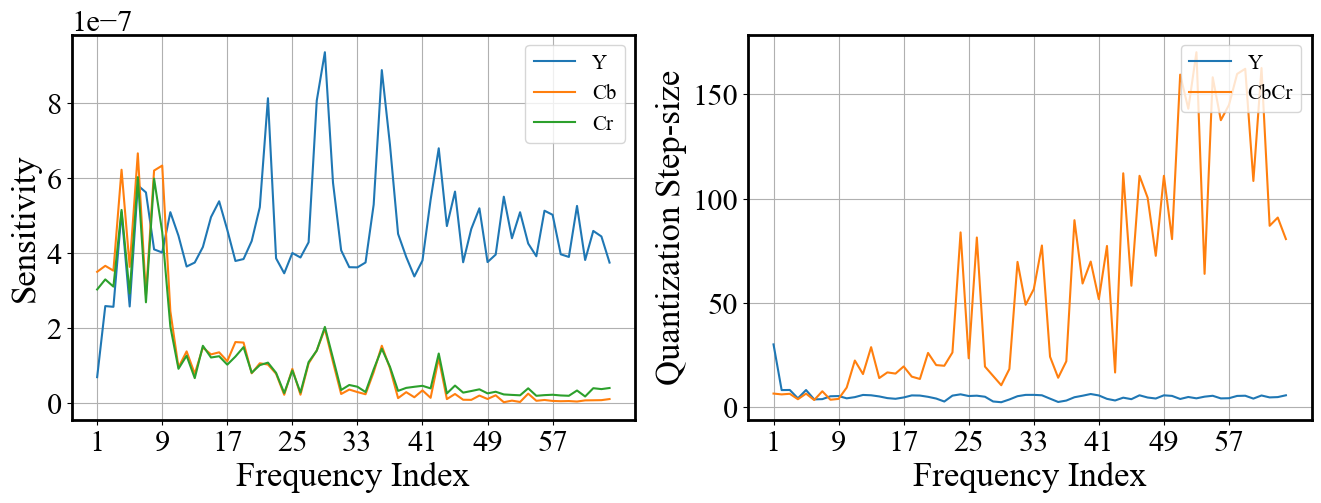}}
    \caption{The sensitivity values (left) and the corresponding sensitivity-based initial quantization tables (right) on ImageNet-1K.}
    \label{fig:sensitivity_imagenet}
\end{figure}

The sensitivity values of the three models used in the CUB200 task are presented in \cref{fig:sensitivity_cub200}. Similarly, \cref{fig:sensitivity_imagenet} illustrates the sensitivity values of the four models used in the ImageNet-1K task, along with their corresponding sensitivity-based initial quantization tables using $\beta = 30$ as an example.

\begin{figure}[!ht]
    \centering
    \subfloat[DeepLabv3-MobileNet]{\includegraphics[width=\linewidth,keepaspectratio]{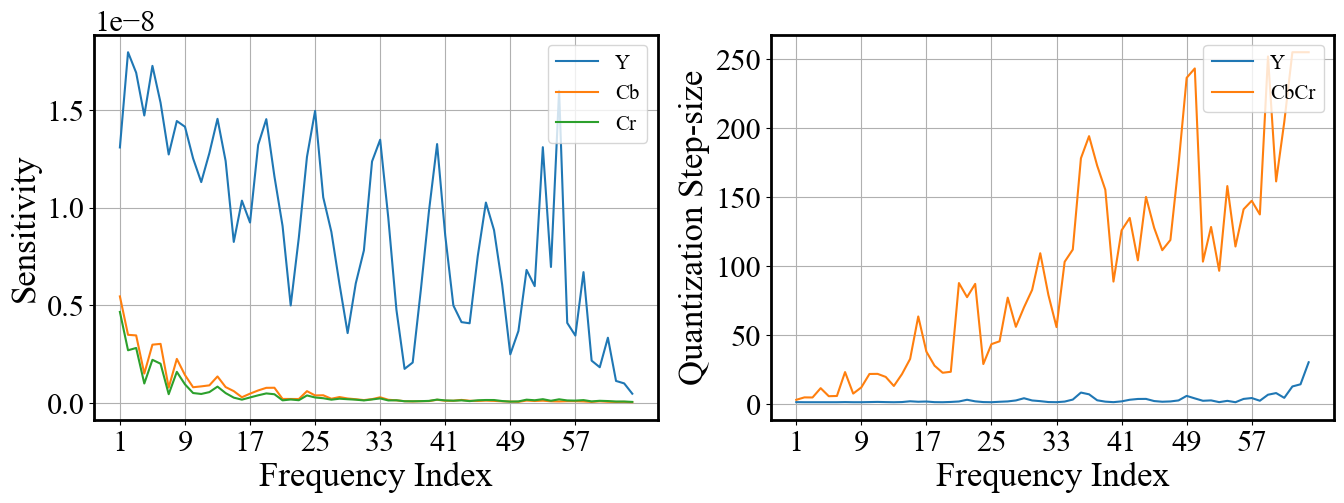}}
    \hfill
    \subfloat[DeepLabv3-ResNet50]{\includegraphics[width=\linewidth,keepaspectratio]{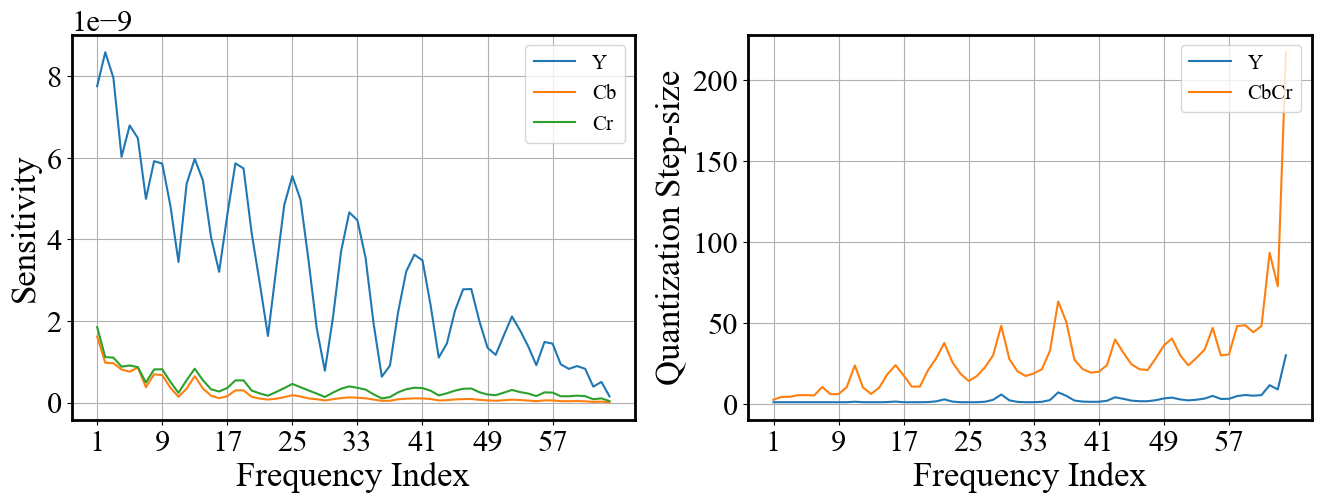}}
    \caption{The sensitivity values (left) and the corresponding sensitivity-based initial quantization tables (right) on Pascal VOC 2012.}
    \label{fig:sensitivity_voc}
\end{figure}

\section{Treatment of Training Complexity Caused by the Differentiable Soft Quantizer} \label{sec:sup_comp}

As mentioned in \cref{sec:Exp_setting}, we configure the reconstruction space $\mathcal{\hat{A}}$ with $L = 1023$, so the total length of the reconstruction space is $2L + 1 = 2047$ (see \cref{eq:recon_space}). Recall that CPMF $P_{\alpha}(\cdot|C)$ is computed using \cref{eq:CPMF_2}. However, the high dimensionality of the reconstruction space introduces significant computational complexity during training. To tackle this issue, we constrain the support of every CPMF to be the Top-5 closest points in $\mathcal{\hat{A}}$ to the coefficient $C$ being quantized. The resulting CPMF is termed the masked CPMF, distinguishing it from the original full-space CPMF. The remainder of this section demonstrates that this simplification has a negligible impact on the effectiveness of $\mathcal{Q}_d$ while successfully addressing the training complexity issue. 

\begin{figure}[!ht]
\centering
\subfloat[$\alpha=1$]{\includegraphics[height=0.20\textwidth]{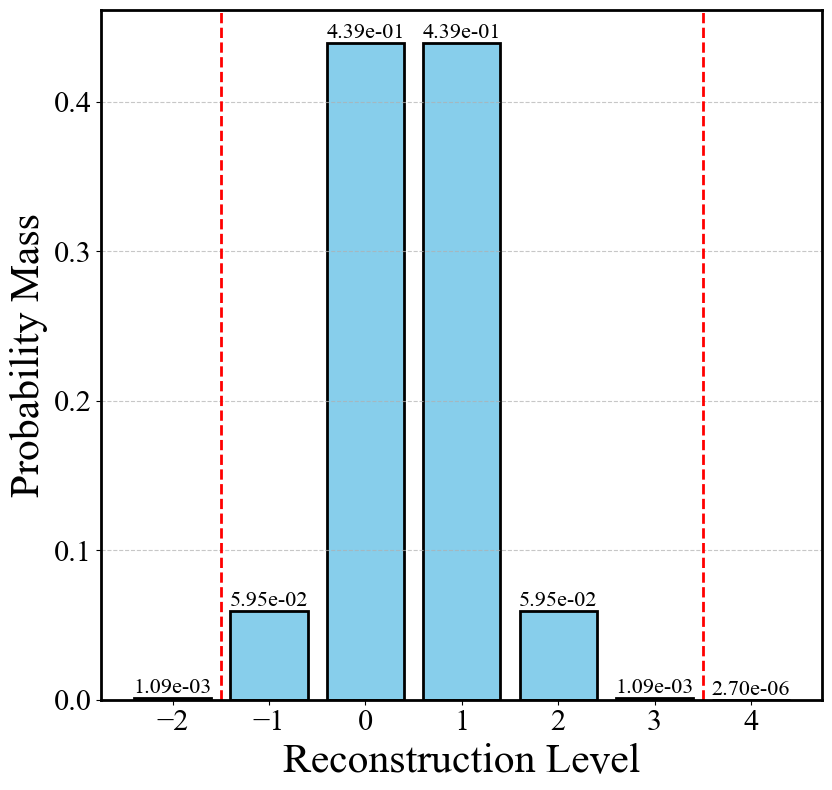}
\label{fig:CPMF-C0.5-alpha1}}
\hfil
\subfloat[$\alpha=100$]{\includegraphics[height=0.20\textwidth]{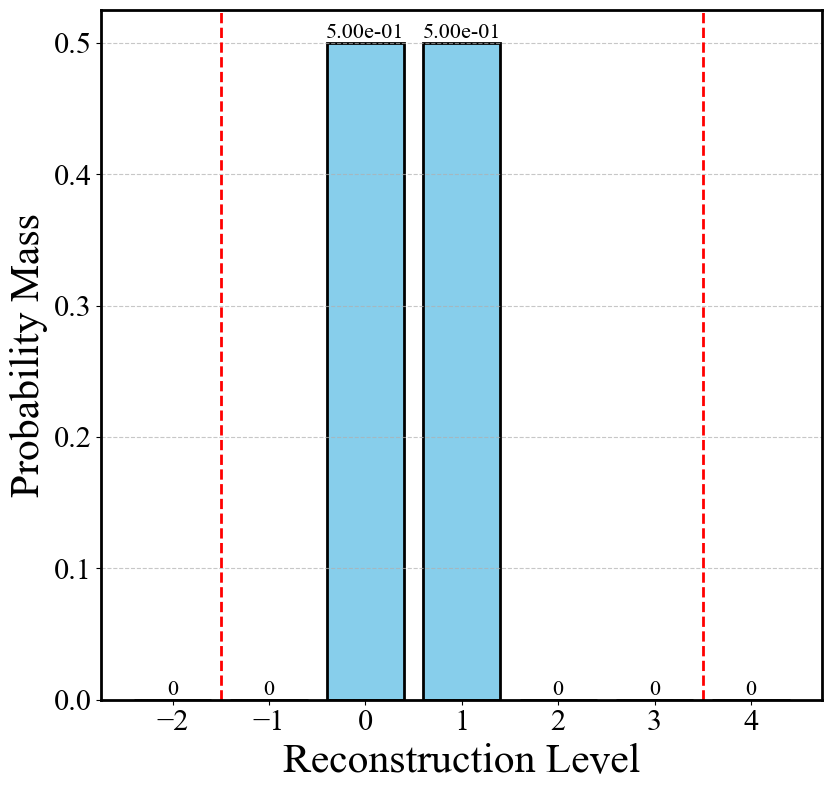}
\label{fig:CPMF-C0.5-alpha100}}
\caption{Partial visualization of CPMFs $P_{\alpha}(\cdot|C=0.5)$ computed by \cref{eq:CPMF_2}, with $L=1023$, $q=1$: (a) $\alpha=1$; and (b) $\alpha=100$. The reconstruction levels between two red dashed lines compose $\mathcal{M}$.}
\label{fig:CPMF}
\end{figure}

\begin{table}[!ht]
\centering
\renewcommand{\arraystretch}{0.8} 
\resizebox{0.55\linewidth}{!}{
\begin{tabular}{l|ccc}
\toprule
\multirow{2}{*}{$\alpha$} & \multicolumn{3}{c}{$\mathcal{Q}_d(C=0.5)$} \\ 
\cmidrule{2-4}
& Full Space & Masked & $\Delta$ \\ 
\midrule
1    & 0.5 & 0.5027 & 0.0027 \\
\cmidrule{1-4}
100  & 0.5 & 0.5 & 0 \\
\bottomrule
\end{tabular}}
\caption{$\mathcal{Q}_d(C=0.5;q=1,\alpha)$ values resulting from the full-space CPMF and the masked CPMF at different $\alpha$, alongside the absolute difference $\Delta$ between the full-space and masked cases.}
\label{tab:CPMF}
\end{table}

Specifically, \cref{fig:CPMF} illustrates how fast the probability mass of a reconstruction level $\hat{C}$ decays in CPMFs as $\hat{C}$ deviates from $\mathcal{Q}_u(C)$. It's clear that the set $\mathcal{M}=\{\mathcal{Q}_u(C)-2q, \mathcal{Q}_u(C)-q, \mathcal{Q}_u(C), \mathcal{Q}_u(C)+q, \mathcal{Q}_u(C)+2q\}$ of reconstruction levels retain a total probability mass close to 1, even in an adversarially chosen scenario where both $q$ and $\alpha$ are extremely small\footnote{As $q$ and $\alpha$ increase, the CPMF becomes even sharper, and as a result, $\mathcal{M}$ will capture increasingly higher probability mass.} (\ie, 1) and the coefficient $C$ being quantized lies exactly at a quantization threshold.  Therefore, instead of computing the full-space CPMF with length $2L+1$ following \cref{eq:CPMF_2}, we only conduct softmax on $\mathcal{M}$ to get the masked CPMF with length 5. The error in $\mathcal{Q}_d$ caused by this masking is negligible as shown in \cref{tab:CPMF}, especially in the high $\alpha$ case. Therefore, it's totally justified to simplify the full-space CPMF to the masked CPMF in our implementation.

On the other hand, we present the training throughput and memory footprint of J4D in \cref{tab:Time_Memory}. Compared to the vanilla training of a DNN model, training quantization tables within the J4D framework incurs only a slight increase in time and space complexity, given the adoption of the masked CPMF.

\begin{table}[!ht]
\centering
\renewcommand{\arraystretch}{0.8} 
\resizebox{\linewidth}{!}{
\begin{tabular}{l|c|c}
\toprule
\multirow{2}{*}{Model (Training Setting)}  & Training            & Peak Memory \\
                                           & Throughput (img/s)  & Usage (MB/GPU) \\
\midrule
Swin-T (Vanilla)                        & 217 & 7106 \\
Swin-T (J4D with Masked CPMFs)          & 188 & 8218 \\
\cmidrule{1-3}
ConvNeXt-tiny (Vanilla)                 & 208  & 9427 \\
ConvNeXt-tiny (J4D with Masked CPMFs)   & 179  & 10725 \\
\bottomrule
\end{tabular}
}
\caption{Comparison of training throughput and memory footprint between different training settings measured with Swin-T and ConvNeXt-tiny on ImageNet-1K, following the configuration specified in \cref{sec:Exp_setting}. The evaluation is conducted on 4 NVIDIA RTX A5000 GPUs. Note that the vanilla training refers to the standard process of training a DNN model (solely for accuracy) using backpropagation, without prepending a trainable JPEG codec.}
\label{tab:Time_Memory}
\end{table}

In conclusion, thanks to the use of the masked CPMF, we effectively control the computational complexity of the J4D training framework while preserving its performance.

\section{Comparison with the Sensitivity-based Method} \label{sec:sup_opts}

\begin{figure}[!t]
    \centering
    \begin{subfigure}{0.23\textwidth}
        \centering
        \includegraphics[width=\linewidth]{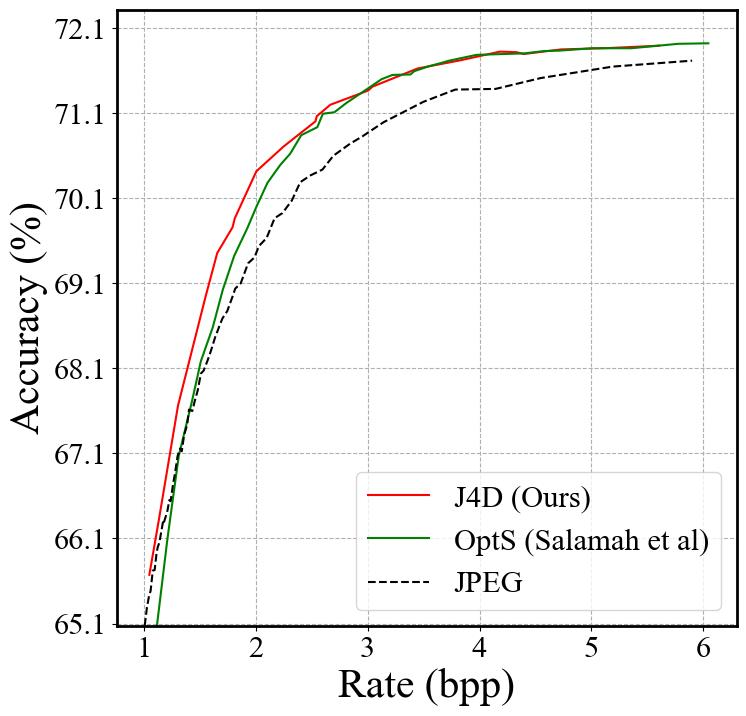}
        \caption{MobileNetV2 (ImageNet-1K)}
        \label{refig:mobilenet_v2_opts}
    \end{subfigure}
    \hfill
    \begin{subfigure}{0.23\textwidth}
        \centering
        \includegraphics[width=\linewidth]{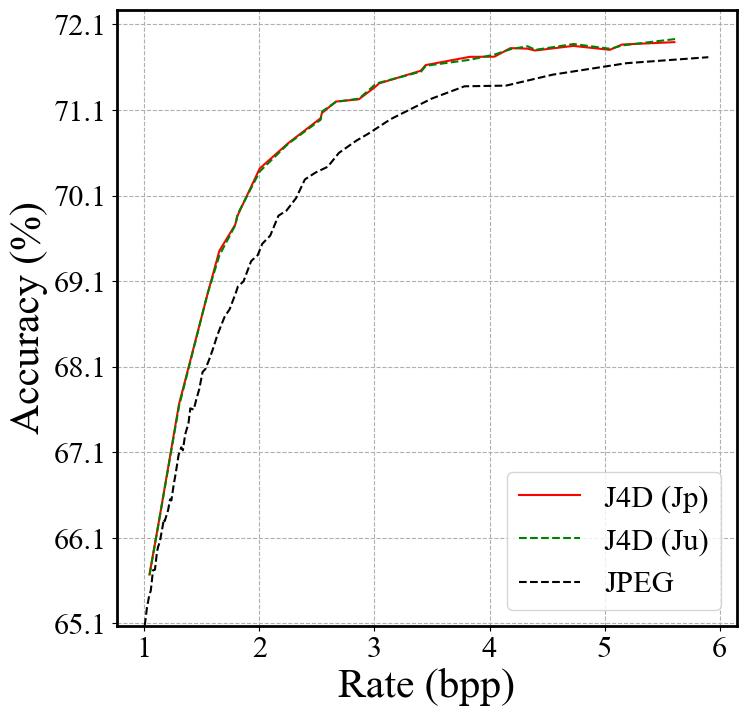}
        \caption{MobileNetV2 (ImageNet-1K)}
        \label{refig:mobilenet_v2_Ju_Jp}
    \end{subfigure}
    
    \vspace{0.3cm}
    
    \begin{subfigure}{0.23\textwidth}
        \centering
        \includegraphics[width=\linewidth]{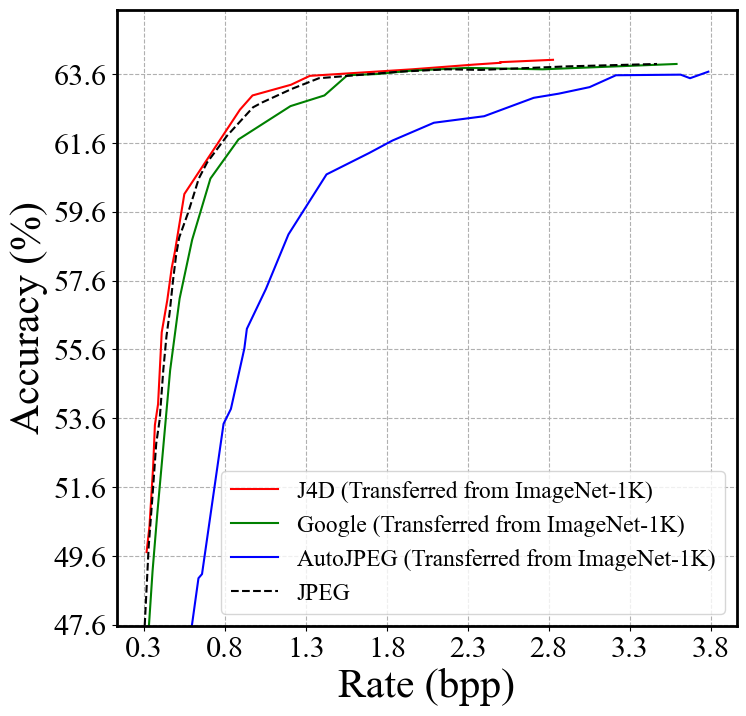}
        \caption{MobileNetV2 (CUB200)}
        \label{refig:mobilenetv2_imagenet_to_cub200}
    \end{subfigure}
    \hfill
    \begin{subfigure}{0.23\textwidth}
        \centering
        \includegraphics[width=\linewidth]{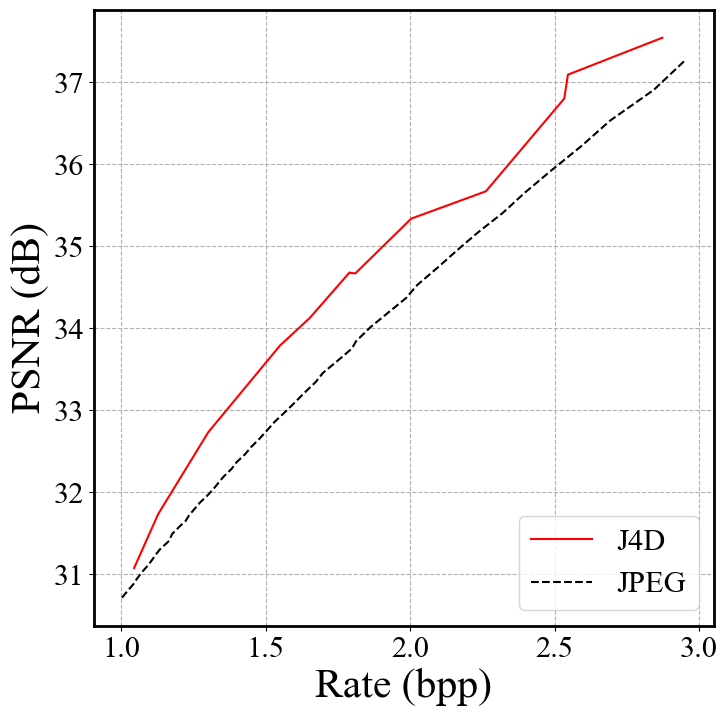}
        \caption{MobileNetV2 (ImageNet-1K)}
        \label{refig:RD-tradeoff}
    \end{subfigure}
    
    \caption{(a) Comparison with a sensitivity-based method dubbed OptS. (b) Comparison of $\mathcal{J}_p$ and $\mathcal{J}_u$ during validation. (c) Evaluation of quantization table transferability from ImageNet-1K to CUB200 given the same DNN architecture. (d) R-D performance evaluation for J4D and JPEG.}
    \label{refig:refigure}
\end{figure}

Using MobileNetV2 on ImageNet-1K as an example, we compare a sensitivity-based method, named OptS \cite{Zheng2023jpegcompliant, Salamah2024jpegcompliant}, with our proposed method, shown in \cref{refig:mobilenet_v2_opts}. The result shows that OptS achieves comparable performance as J4D at high rate; however, in the low rate case, OptS is much worse than J4D and ultimately becomes worse than the default JPEG as the rate approaches 1 bpp. This is consistent with our discussion in \cref{sec:Related_work}, where we mention that OptS, or sensitivity-based methods in general, may not do well in the low rate case due to the reliance on a first-order Taylor approximation of DNN loss functions.

\section{Using Uniform Quantization During the Actual JPEG Encoding Phase} \label{sec:jp_ju}

For consistency and coherence of the proposed method, we choose to use the probabilistic quantization during actual JPEG encoding, \ie, $\mathcal{J}_p$, for the validation set in the main paper. However, once quantization tables are learned, to facilitate deployment within the conventional JPEG pipeline, $\mathcal{J}_p$ can be replaced with the uniform quantizer $\mathcal{J}_u$ parameterized by the learned quantization tables. As shown in \cref{refig:mobilenet_v2_Ju_Jp}, the performances of $\mathcal{J}_u$ and $\mathcal{J}_p$, both with the learned quantization tables, are almost identical since $\alpha=100$ makes $Q_p$ equivalent to $Q_u$ with high probability. This confirms that the quantization tables learned using J4D can be seamlessly integrated into the conventional JPEG-based systems, since $\mathcal{J}_u$ can be used in place of $\mathcal{J}_p$ without degrading R-A performance.

\section{Quantization Table Transferability Across Datasets} \label{sec:data_trans}

In the main body of the paper, we evaluate the transferability of optimized quantization tables among DNN models on the same dataset. In this section, we further extend our investigation to quantization table transferability across different datasets. In particular, we transfer the optimized quantization tables for MobileNetV2 on ImageNet-1K to MobileNetV2 on CUB200 and evaluate the resulting R-A performance. As shown in \cref{refig:mobilenetv2_imagenet_to_cub200}, quantization tables obtained by J4D demonstrate decent dataset transferability, outperforming the default JPEG, whereas those obtained by Google and AutoJPEG lead to suboptimal performance when transferred from ImageNet-1K to CUB200.

\section{Rate-Distortion Performance} \label{sec:r_d}

Although the main target of J4D is to learn quantization tables suitable for DNN vision, we also analyze its rate-distortion (R-D) performance for human vision. As shown in \cref{refig:RD-tradeoff}, the R-D performance of quantization tables optimized for MobileNetV2 on ImageNet-1K is better than that of the default JPEG. This highlights a promising property of J4D that it not only improves the image quality for DNN vision, but also improves the image quality in human eyes as a beneficial by-product.

\section{Additional Discussion on Related Work} \label{sec:supp_related_work}

Another related work \cite{choi2020task} also adopts a backpropagation-based training framework to address JPEG compression for DNN vision. However, our approach is fundamentally different and based on information theory. Specifically, the method in \cite{choi2020task} relies on (1) two auxiliary DNNs on the JPEG encoding side to provide image-adaptive quantization tables and (2) three DNNs to estimate the rate. In contrast, no auxiliary DNN is required in our case; once the learned quantization tables are obtained, there is no change to the JPEG encoding side. Unfortunately, due to the absence of publicly available source code or pre-trained models from \cite{choi2020task}, we are unable to reproduce their results for direct benchmarking.

\end{document}